\documentclass[journal]{IEEEtran}
\usepackage{amssymb,amsthm,amsmath}
\usepackage{amsmath}
\usepackage{multirow}
\usepackage{algorithm}
\usepackage{algorithmicx}
\usepackage{algpseudocode}
\usepackage{lipsum}
\usepackage{graphicx}
\usepackage{multicol}
\usepackage{float}

\begin{document}
%
\title{A Robust Alternating Direction Method for Constrained Hybrid Variational Deblurring Model}
\author{Ryan~Wen~Liu$^{*}$
        and~Tian~Xu
\thanks{$^{*}$Corresponding author. This work was supported in part by the NSFC Fund (Grant 51179147). R. W. Liu is with the Department of Mathematics, Wuhan University of Technology, Wuhan 430070, China (e-mail: lwsunlight@gmail.com). T. Xu is with the Department of Communication Engineering, Wuhan University, Wuhan 430072, China (e-mail: xutian9@gmail.com).}}
\markboth{DRAFT}%
{Shell \MakeLowercase{\textit{Liu et al.}}: Constrained Hybrid Variational Deblurring Model}
\maketitle
%
\begin{abstract}
In this work, a new constrained hybrid variational deblurring model is developed by combining the non-convex first- and second-order total variation regularizers. Moreover, a box constraint is imposed on the proposed model to guarantee high deblurring performance. The developed constrained hybrid variational model could achieve a good balance between preserving image details and alleviating ringing artifacts. In what follows, we present the corresponding numerical solution by employing an iteratively reweighted algorithm based on alternating direction method of multipliers. The experimental results demonstrate the superior performance of the proposed method in terms of quantitative and qualitative image quality assessments.
\end{abstract}
%
\IEEEpeerreviewmaketitle
\section{Introduction}
%
%
\IEEEPARstart{I}{mage} deblurring is a well-known ill-posed inverse problem, which has attracted increasing attention from many sectors. In this work, we proceed to work with the matrix-vector representation of the image degradation process, i.e.,
\begin{equation}
\mathbf{g} = \mathbf{H} \mathbf{f} + \epsilon,
\end{equation}
where $\mathbf{f} \in \mathcal{R}^{mn}$ denotes an original image of size $m \times n$, $\mathbf{g} \in \mathcal{R}^{mn}$ represents the degraded image, $\epsilon \in \mathcal{R}^{mn}$ is the additive Gaussian noise, and $\mathbf{H} \in \mathcal{R}^{mn \times mn}$ related to the boundary conditions denotes a blurring matrix. To cope with the ill-posed nature of deblurring problem, a large number of regularization techniques have been developed. Probably (iterated) Tikhonov regularization and its variants \cite{DonatelliHanke},\cite{LiuWu} are the most popular regularization methods. However, they tend to over-smooth image details. In order to overcome this drawback, total variation (TV) based regularization method was proposed \cite{Rudin}, which could preserve edges and discontinuities due to its nature in favoring piecewise constant solution.
%
%

In practice, the undesired staircase effect is often present in the first-order TV-based deblurring results \cite{Rudin}. To reduce the staircase effect, an increasing effort has been paid to replace the classical TV norm \cite{Rudin} by the second-order TV norm \cite{LvSongWang},\cite{Benning}. Nevertheless, the second-order version could lead to poor edge-preserving performance. Naturally, a convex combination of the first- and second-order TV regularizers can control the tradeoff between artifact-suppression and edge-preservation \cite{PapafitsorosSchonlieb}. Recent research in natural image statistics illustrates the gradients can be well distributed as the heavy-tailed hyper-Laplacian distribution $( p(x) \propto e^{-\tau \left| x \right|^{\pi}} )$ with $0.5 \le \pi \le 0.8$ \cite{Krishnan}. Based on maximum a posteriori (MAP), non-convex first-order TV \cite{Krishnan} has been proposed for image deblurring. From a statistical point of view, the second-order derivatives also follow the hyper-Laplacian distribution. Theoretically and practically, the MAP-based non-convex second-order regularizer is superior to the convex formulation in terms of edge preservation \cite{Oh}. Motivated by the works of \cite{Krishnan},\cite{Oh}, it is natural to investigate a new hybrid variational deblurring model, which can take advantages of the two non-convex TV regularizers and overcome their shortcomings.

However, the pixel intensity values generated by TV-based models usually move out of a given dynamic range $\left[ l, u \right]$, for instance, $\left[0, 1 \right]$ for normalized images and $\left[0, 255 \right]$ for 8-bit images. In this condition, a projection operation is necessary to map the ``outliers" back into the dynamic range. Nevertheless, it becomes difficult to guarantee that the projected image is the minimizer of the TV-based models \cite{BeckTeboulle},\cite{ChanTao}. To further improve the deblurring performance, a box constraint $\left[ l, u \right]$ is embedded into the proposed hybrid variational model in this paper. It is well known that the TV-based models are always computationally hard to solve owning to the high nonlinearity and non-differentiability of the TV term. In order to effectively and robustly solve the proposed deblurring model, we present an iteratively reweighted algorithm based on alternating direction method of multipliers (ADMM) \cite{ChanKhoshabeh}.
%
%
%
\subsection{Contributions}
In this paper, we propose a constrained hybrid variational model for non-blind deblurring. The proposed method significantly differs from previous works in the following aspects:

\begin{enumerate}
\item The constrained hybrid variational model, which combines the advantages of the non-convex first- and second-order TV regularizers, could effectively preserve image details while suppressing ringing and noise artifacts.
\item In the constrained deblurring framework, we develop an ADMM-based iteratively reweighted algorithm to solve the non-convex minimization problem whose subproblems have their own closed form solutions.
\end{enumerate}

The accuracy and robustness of the proposed deblurring model will be verified by a series of numerical experiments.

\section{Proposed Scheme}
\subsection{Constrained Hybrid Variational Deblurring Model}
To suppress ringing artifacts while preserving image details, a hybrid deblurring model is proposed by combining the non-convex first- and second-order TV regularizers, i.e.,
\begin{align}\label{hybrid}
\min_{\mathbf{f}} \Big\{ {\frac{\mu}{2}\left\| \mathbf{Hf - g} \right\|^{2}_{2}} &+ \zeta {\left\| \mathbf{Df} \right\|}^{\nu_{1}}_{1} + \left(1 - \zeta \right){\left\| \mathbf{D}^{2} \mathbf{f} \right\|}^{\nu_{2}}_{1} \Big\},
\end{align}
where $\mu > 0$ is a regularization parameter, $\nu_{1}$ and $\nu_{2}$ are related to the hyper-Laplacian distributions of the first- and second-order derivatives, $\mathbf{D}$ and $\mathbf{D}^{2}$ denote the finite-difference operators of the first- and second-order, respectively. The adaptive weighting function $\zeta \in [0,1]$ maintains a balance between artifact reduction and detail preservation. To preserve image details in texture and edge regions, the function $\zeta$ should be close to $1$. In contrast, the $\zeta$ should be small and almost close to $0$ in homogeneous regions to suppress ringing artifacts. In this work, the $\zeta$ is achieved based on eigenvalues of the Hessian matrix of each pixel in an image \cite{Tian}. Given an image $\mathbf{f}_{k}$ at the $k$-th iteration in (\ref{hybrid}), we use a Gaussian filtered version of the Hessian matrix $J_{\sigma}(\mathbf{f}_{k})$ to improve calculation robustness in noisy conditions.
%
\begin{equation*}
J_{\sigma} (\mathbf{f}_{k}) =
\begin{bmatrix}
\mathbf{D}_{xx}(G_{\sigma} * \mathbf{f}_{k}) & \mathbf{D}_{xy}(G_{\sigma} * \mathbf{f}_{k}) \\
\mathbf{D}_{yx}(G_{\sigma} * \mathbf{f}_{k}) & \mathbf{D}_{yy}(G_{\sigma} * \mathbf{f}_{k})
\end{bmatrix},
\end{equation*}
where $*$ is the convolution operator, and $G_{\sigma}$ denotes the Gaussian kernel function. Let $\lambda_{1}$ and $\lambda_{2}$ denote two distinct eigenvalues of the matrix $J_{\sigma}(\mathbf{f}_{k})$. Here the larger eigenvalue $\lambda_{1}$ and smaller one $\lambda_{2}$ correspond to the maximum and minimum local variation at a pixel $(x,y) \in \Omega$ (image domain), respectively. The weighting function $\zeta_{k}(x,y)$ at the $k$-th iteration is then given by
\begin{equation}\label{weighting}
\zeta_{k} (x,y) = 1 - \frac{1}{1 + \kappa \left( \lambda_{1} - \lambda_{2} \right) \cdot \frac{\varrho_{k} (x,y) - \mathrm{min} (\varrho_{k})} {\mathrm{max}(\varrho_{k}) - \mathrm{min} (\varrho_{k})}},
\end{equation}
where $\kappa$ is a constant, and $\varrho_{k}(x,y)$ represents the local gray-level variance of image $\mathbf{f}_{k}$. Let $\Omega^{\tilde{\omega}}(x,y) \subseteq \Omega$ denote the set of pixel-coordinates in a $\tilde{\omega}\text{-by-}\tilde{\omega}$ region centered at $(x,y) \in \Omega^{\tilde{\omega}}(x,y)$, then the local variance $\varrho_{k}(x,y)$ can be calculated as follows:
\begin{equation*}
\varrho_{k}(x,y) = \frac{1}{\tilde{\omega} \times \tilde{\omega}}
\begin{matrix}
\sum_{(\tilde{x},\tilde{y}) \in \Omega^{\tilde{\omega}}(x,y)} \left\| \mathbf{f}_{k}(\tilde{x},\tilde{y}) - \mathbf{f}_{k} (x,y) \right\|_{2}^{2},
\end{matrix}
\end{equation*}
where $\Omega^{\tilde{\omega}}(x,y) = \left\{ (x + \hat{x}, y + \hat{y}) : -\frac{\tilde{\omega} - 1}{2} \leq \hat{x},\hat{y} \leq \frac{\tilde{\omega} - 1}{2} \right\}$, and the $\tilde{\omega}$ is a positive odd integer. In this paper, the size of $\Omega^{\tilde{\omega}}(x,y)$ is set to $5 \text{-by-} 5$. However, the restored values from model (\ref{hybrid}) usually move out of a given dynamic range $\left[ 0, 1 \right]$ or $\left[ 0, 255 \right]$. In this condition, a projection process should be implemented to bring the values back into the dynamic range. This will bring negative effects on final deblurring performance because the restored values are no longer the minimizer of the TV-based model (\ref{hybrid}). To further improve the deblurring performance, a box constraint $\left[ l, u \right]$ is embedded into the deblurring model (\ref{hybrid}). As a result, we are mainly interested in the following constrained non-convex hybrid TV (CNCHTV) model
\begin{align}\label{Constrainedhybrid}
\min_{\mathbf{f} \in \Phi} \Big\{ {\frac{\mu}{2}\left\| \mathbf{Hf - g} \right\|^{2}_{2}} &+ \zeta {\left\| \mathbf{Df} \right\|}^{\nu_{1}}_{1} + \left(1 - \zeta \right){\left\| \mathbf{D}^{2} \mathbf{f} \right\|}^{\nu_{2}}_{1} \Big\},
\end{align}
where $\Phi = \left\{ \mathbf{f} \in \mathcal{R}^{mn} ~|~ l \le \mathbf{f} \le u\right\}$ is a convex closed set. In practice, we set the box constraint $\left[ 0,255 \right]$ in model (\ref{Constrainedhybrid}).
\subsection{An ADMM-Based Iteratively Reweighted Algorithm}
To effectively solve the non-convex hybrid deblurring model (\ref{Constrainedhybrid}), the convex approximation of (\ref{Constrainedhybrid}) at each iteration can be achieved as follows
\small
\begin{equation}\label{convexequation}
\mathbf{f}_{k+1} = \min_{\mathbf{f} \in \Phi} \Big\{ {\frac{\mu}{2}\left\| \mathbf{Hf - g} \right\|^{2}_{2}} + \zeta_{k} \psi_{1}^{k} {\left\| \mathbf{Df} \right\|}_{1} + \left(1 - \zeta_{k} \right) \psi_{2}^{k} {\left\| \mathbf{D}^{2} \mathbf{f} \right\|}_{1} \Big\}
\end{equation}
\normalsize
where variables $\psi_{1}^{k} = {\left\| \mathbf{D} \mathbf{f}_{k} \right\|}_{1}^{\nu_{1} - 1}$ and $\psi_{2}^{k} = {\left\| \mathbf{D}^{2} \mathbf{f}_{k} \right\|}_{1}^{\nu_{2} - 1}$. For the sake of better reading, we omit the index $k$ for $\zeta_{k}$, $\psi_{1}^{k}$ and $\psi_{2}^{k}$. We first introduce three intermediate variables $\mathbf{v}$, $\mathbf{w}$ and $\mathbf{u}$, then use the ADMM to solve (\ref{convexequation}), i.e.,
\small
\begin{equation}\label{equivalent}
\begin{split}
&\min_{\mathbf{v}, \mathbf{w}, \mathbf{u} \in \Phi, \mathbf{f}} \left\{ {\frac{\mu}{2}\left\| \mathbf{Hf - g} \right\|^{2}_{2}} + \zeta \psi_{1} {\left\| \mathbf{v} \right\|}_{1} + \left(1 - \zeta \right) \psi_{2} {\left\| \mathbf{w} \right\|}_{1} \right\}\\
&~~~~\mathrm{s.t.} ~~~~ \mathbf{v} = \mathbf{Df}, ~ \mathbf{w} = \mathbf{D}^{2} \mathbf{f}, ~ \mathbf{u} = \mathbf{f}.
\end{split}
\end{equation}
\normalsize

Let $\mathcal{L}_{\mathcal{A}}(\mathbf{v}, \mathbf{w}, \mathbf{u}, \mathbf{f}; \omega, \lambda, \xi)$ be the augmented Lagrangian function of (\ref{equivalent}) which is defined as follows
\small
\begin{equation}\label{Lagrangian}
\begin{split}
\mathcal{L}_{\mathcal{A}}&( \mathbf{v}, \mathbf{w}, \mathbf{u}, \mathbf{f}; \omega, \lambda, \xi) \\
& = \frac{\mu}{2}\left\| \mathbf{Hf - g} \right\|^{2}_{2} + \zeta \psi_{1} {\left\| \mathbf{v} \right\|}_{1} + \frac{\beta_{1}}{2}\left\| \mathbf{v} - \mathbf{Df} \right\|^{2}_{2}\\
& - \omega^{T} \left( \mathbf{v} - \mathbf{Df}\right) + \left(1 - \zeta \right) \psi_{2} {\left\| \mathbf{w} \right\|}_{1} + \frac{\beta_{2}}{2}\left\| \mathbf{w} - \mathbf{D}^{2} \mathbf{f} \right\|^{2}_{2} \\
& - trace\left( \lambda^{T} \left( \mathbf{w} - \mathbf{D}^{2} \mathbf{f} \right) \right) - \xi^{T} \left( \mathbf{u} - \mathbf{f} \right) + \frac{\beta_{3}}{2} \left\| \mathbf{u} - \mathbf{f} \right\|^{2}_{2}
\end{split},
\end{equation}
\normalsize
where $\beta_{1},\beta_{2},\beta_{3} > 0$ are penalty parameters, and $\omega \in \mathcal{R}^{2mn}$, $\lambda \in \mathcal{R}^{4mn}$ and $\xi \in \mathcal{R}^{mn}$ are the Lagrange multiplies. We alternatively solve (\ref{Lagrangian}) with respect to $\mathbf{v}, \mathbf{w}$, $\mathbf{f}$ and $\mathbf{u}$ and then update $\omega$, $\lambda$ and $\xi$. We now investigate these subproblems one by one.

1) $~\mathbf{v}$-subproblem: Since the unknown variable $\mathbf{v}$ is componentwise separable in the subproblem $\mathbf{v}_{k+1} \leftarrow \min_{\mathbf{v}} \mathcal{L}_{\mathcal{A}} (\mathbf{v}, \mathbf{w}_{k}, \mathbf{u}_{k}, \mathbf{f}_{k}; \omega_{k}, \lambda_{k}, \xi_{k})$ in (\ref{Lagrangian}), this subproblem can be effectively solved using the shrinkage operation \cite{ChanTao}. In particular, minimization of the augmented Lagrange function $\mathcal{L}_{\mathcal{A}}$ with respect to $\mathbf{v}$ is equivalent to
\begin{equation*}
\mathbf{v}_{k+1} = \min_{\mathbf{v}} \left\{ \frac{\beta_{1}}{2} \left\| \mathbf{v} - \left( \mathbf{Df}_{k} + \frac{\omega_{k}}{\beta_{1}} \right) \right\|_{2}^{2} + \zeta_{k} \psi_{1} {\left\| \mathbf{v} \right\|}_{1} \right\}.
\end{equation*}
Let $\chi_{k}^{\mathbf{v}} = \mathbf{D} \mathbf{f}_{k} + {\omega_{k}}/{\beta_{1}}$, $\mathbf{v}_{k+1}$ is then given by
\begin{equation}\label{vsubproblem}
\begin{split}
\mathbf{v}_{k+1} & =  \mathrm{shrink} \left( \chi_{k}^{\mathbf{v}}, \frac{\zeta_{k} \psi_{1}}{\beta_{1}} \right)\\
& = \mathrm{max} \left\{ \left\| \chi_{k}^{\mathbf{v}} \right\|_{2} - \frac{\zeta_{k} \psi_{1}}{\beta_{1}} , 0 \right\} \circ \mathrm{sgn} \left( \chi_{k}^{\mathbf{v}} \right)
\end{split},
\end{equation}
here $\circ$ and $\mathrm{sgn}$ are the point-wise product and signum function, respectively.

2) $~\mathbf{w}$-subproblem: Similarly, let $\chi_{k}^{\mathbf{w}} = \mathbf{D}^{2} \mathbf{f}_{k} + {\lambda_{k}} / {\beta_{2}}$, then the solution of the $\mathbf{w}$-subproblem $\mathbf{w}_{k+1} \leftarrow \min_{\mathbf{w}} \mathcal{L}_{\mathcal{A}}(\mathbf{v}_{k+1}, \mathbf{w}, \mathbf{u}_{k}, \mathbf{f}_{k}; \omega_{k}, \lambda_{k}, \xi_{k})$ in (\ref{Lagrangian}) is given by
%
%
\begin{equation}\label{wsubproblem}
\mathbf{w}_{k+1} = \mathrm{max} \left\{ \left\| \chi_{k}^{\mathbf{w}} \right\|_{2} - \frac{\left(1 - \zeta_{k} \right) \psi_{2}}  {\beta_{2}}, 0\right\} \circ \mathrm{sgn} \left( \chi_{k}^{\mathbf{w}} \right).
\end{equation}

3) $~\mathbf{u}$-subproblem: The solution of the subproblem $\mathbf{u}_{k+1} \leftarrow \min_{\mathbf{u}} \mathcal{L}_{\mathcal{A}}(\mathbf{v}_{k+1}, \mathbf{w}_{k+1}, \mathbf{u}, \mathbf{f}_{k}; \omega_{k}, \lambda_{k}, \xi_{k})$ can be implemented by a simple projection $\mathcal{P}_{\Phi}$ onto the box constraint $\left[0,255\right]$, i.e.,
\begin{equation}\label{usubproblem}
\mathbf{u}_{k+1} = \mathcal{P}_{\Phi} \left( \mathbf{f}_{k} + \frac{\xi_{k}}{\beta_{3}} \right).
\end{equation}

4) $~\mathbf{f}$-subproblem: The solution of the subproblem $\mathbf{f}_{k+1} \leftarrow \min_{\mathbf{f}} \mathcal{L}_{\mathcal{A}}(\mathbf{v}_{k+1}, \mathbf{w}_{k+1}, \mathbf{u}_{k+1}, \mathbf{f}; \omega_{k}, \lambda_{k}, \xi_{k})$ can be obtained by considering the following normal equation
%
\begin{equation}\label{fsubproblem_A}
\begin{split}
& \left( \mu \mathbf{H}^{T} \mathbf{H} + \beta_{1} \mathbf{D}^{T} \mathbf{D} + \beta_{2} \left(\mathbf{D}^{2}\right)^{T} \mathbf{D}^{2} + \beta_{3} \mathbf{I} \right) \mathbf{f} \\
& ~~~~~~ = \mu \mathbf{H}^{T} \mathbf{g} + \beta_{1} \mathbf{D}^{T} \left( \mathbf{v}_{k+1} - \frac{\omega_{k}}{\beta_{1}}\right) \\
& ~~~~~~ + \beta_{2} \left(\mathbf{D}^{2}\right)^{T} \left( \mathbf{w}_{k+1} - \frac{\lambda_{k}}{\beta_{2}}\right) + \beta_{3} \left( \mathbf{u}_{k+1} - \frac{\xi_{k}}{\beta_{3}} \right)
\end{split}
\end{equation}
%

Under the periodic boundary condition for $\mathbf{f}$, $\mathbf{D}^{T} \mathbf{D}$, $\left(\mathbf{D}^{2}\right)^{T} \mathbf{D}^{2}$ and $\mathbf{H}^{T} \mathbf{H}$ are all block circulant matrices with circulant blocks and thus are diagonalizable by the 2D discrete Fourier transforms (DFTs) \cite{ChanTao}. Consequently, the equation (\ref{fsubproblem_A}) can be solved by one forward DFT and one inverse DFT.

5) $~\omega$, $\lambda$ and $\xi$ update: We update the Lagrange multiplies $\omega$, $\lambda$ and $\xi$ as follows
%
\begin{align}
        \omega_{k+1}  & = \omega_{k} - \gamma \beta_{1} \left( \mathbf{v}_{k+1} - \mathbf{D} \mathbf{f}_{k+1} \right) \label{updateomega} \\
        \lambda_{k+1} & = \lambda_{k} - \gamma \beta_{2} \left( \mathbf{w}_{k+1} - \mathbf{D}^{2} \mathbf{f}_{k+1} \right) \label{updatelambda} \\
        \xi_{k+1}     & = \xi_{k} - \gamma \beta_{3} \left( \mathbf{u}_{k+1} - \mathbf{f}_{k+1} \right) \label{updatexi}
\end{align}
where steplength $\gamma = 1.618$ is adopted in (\ref{updateomega}-\ref{updatexi}). Algorithm 1 shows the pseudocode of the robust alternating direction method for CNCHTV deblurring model (\ref{convexequation}).
\begin{algorithm}[H]
    \caption{An ADMM-based iteratively reweighted algorithm for CNCHTV minimization problem.}
    \begin{algorithmic}
    \State \textbf{Input:} blurred image $\mathbf{g}$, blurring matrix $\mathbf{H}$ and parameters $\mu$, $\beta_{1}$, $\beta_{2}$, $\beta_{3}$, $\nu_{1}$, $\nu_{2}$.
    \State \textbf{Initialize:} $\mathbf{f}_{0} = \mathbf{u}_{0} = \mathbf{g}$, $\mathbf{v}_{0} = \mathbf{D} \mathbf{f}_{0}$, $\mathbf{w}_{0} = \mathbf{D}^{2} \mathbf{f}_{0}$, $\omega_{0} = 0$, $\lambda_{0} = 0$ and $\xi_{0} = 0$.
    \While{ $a~stopping~criterion~is~not~satisfied$ }
    \State $1.$ Compute $\mathbf{v}_{k+1}$ according to (\ref{vsubproblem}).
    \State $2.$ Compute $\mathbf{w}_{k+1}$ according to (\ref{wsubproblem}).
    \State $3.$ Compute $\mathbf{u}_{k+1}$ according to (\ref{usubproblem}).
    \State $4.$ Compute $\mathbf{f}_{k+1}$ by solving (\ref{fsubproblem_A}).
    \State $5.$ Update Lagrange multipliers $\omega_{k+1}$, $\lambda_{k+1}$ and $\xi_{k+1}$.
    \State $6.$ Update adaptive weights $\psi_{1}^{k+1}$, $\psi_{2}^{k+1}$ and $\zeta_{k+1}$.
    %
    \EndWhile
  \end{algorithmic}
  \label{algorithm1}
\end{algorithm}
\begin{table*}
  \centering
  \caption{Performance Comparison of MSSIM Index on Three Color Images (FMI-ADI-RSI) in the presence of Different Spatially-Invariant PSFs and Gaussian Noise Levels}
  \begin{tabular}{|c|c|c|c|c|c|}
  \hline
  Kernel & Noise level & Blurry + Noisy & Krishnan's method \cite{Krishnan} & Chan's method \cite{ChanKhoshabeh} & CNCHTV model\\
  \hline \hline
    \multirow{4}{*}{$\sharp 1$} & 0\% &0.8381-0.8126-0.6834& 0.9249-0.8999-0.8764 & 0.9522-0.9471-0.9266 &
                                \textbf{0.9846}-\textbf{0.9701}-\textbf{0.9441}\\
                                & 1\% &0.8086-0.7926-0.6805& 0.8905-0.8640-0.8407 & 0.8985-0.8737-0.8611 & \textbf{0.9203}-\textbf{0.9096}-\textbf{0.9083}\\
                                & 2\% &0.7363-0.7403-0.6772& 0.8642-0.8404-0.8152 & 0.8838-0.8531-0.8314 & \textbf{0.8980}-\textbf{0.8824}-\textbf{0.8781}\\
                                & 5\% &0.5078-0.5458-0.6305& 0.8115-0.7853-0.7699 & 0.8287-0.7979-0.7900 & \textbf{0.8495}-\textbf{0.8301}-\textbf{0.8402}\\
                                \hline
    \multirow{4}{*}{$\sharp 2$} & 0\% &0.7969-0.7719-0.5921& 0.9611-0.9423-0.9317 & 0.9889-0.9427-0.9035 &
                                \textbf{0.9946}-\textbf{0.9826}-\textbf{0.9636}\\
                                & 1\% &0.7677-0.7522-0.5896& 0.8879-0.8662-0.8688 & 0.9032-0.8770-0.8874 &
                                \textbf{0.9102}-\textbf{0.9033}-\textbf{0.8994}\\
                                & 2\% &0.6975-0.7011-0.5829& 0.8493-0.8241-0.8119 & 0.8773-0.8470-0.8390 &
                                \textbf{0.8842}-\textbf{0.8672}-\textbf{0.8470}\\
                                & 5\% &0.4739-0.5133-0.5457& 0.7750-0.7537-0.7358 & 0.8141-0.7872-0.7645 &
                                \textbf{0.8392}-\textbf{0.8168}-\textbf{0.8071}\\
                                \hline
    \multirow{4}{*}{$\sharp 3$} & 0\% &0.7873-0.7827-0.5592& 0.9495-0.9204-0.9057 & 0.9768-0.9457-0.9089 &
                                \textbf{0.9948}-\textbf{0.9644}-\textbf{0.9506}\\
                                & 1\% &0.7584-0.7628-0.5569& 0.8898-0.8614-0.8556 & 0.9051-0.8737-0.8799 &
                                \textbf{0.9114}-\textbf{0.9066}-\textbf{0.9065}\\
                                & 2\% &0.6882-0.7119-0.5502& 0.8548-0.8275-0.8086 & 0.8757-0.8439-0.8359 &
                                \textbf{0.8860}-\textbf{0.8758}-\textbf{0.8524}\\
                                & 5\% &0.4655-0.5221-0.5152& 0.7928-0.7716-0.7328 & 0.8098-0.7782-0.7667 &
                                \textbf{0.8384}-\textbf{0.8204}-\textbf{0.8060}\\
  \hline
  \end{tabular}
\end{table*}
%
%
%
\begin{figure}[!h]
      \centering
      \begin{minipage}{.27\linewidth}
            \includegraphics[width=\linewidth]{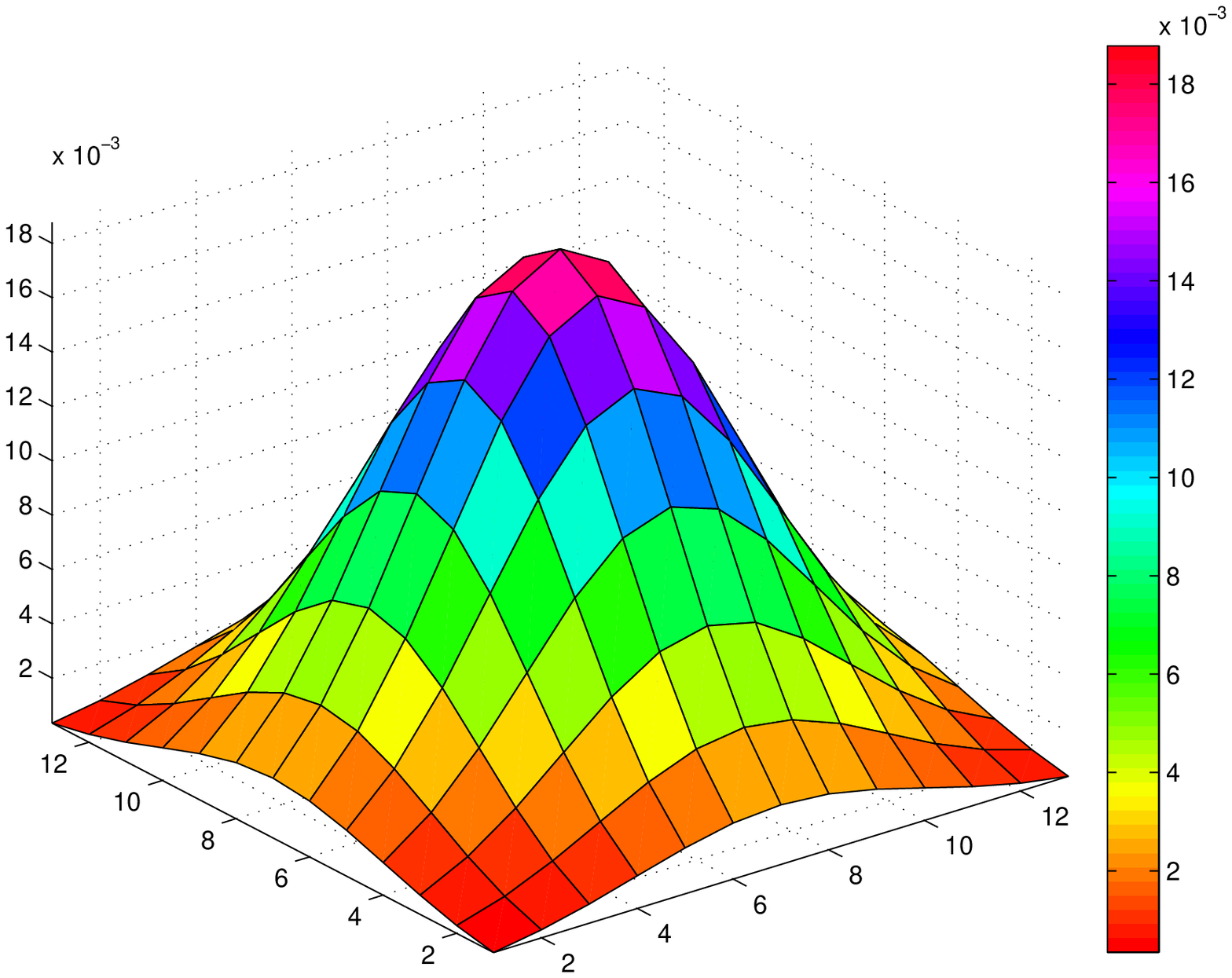}
      \end{minipage}
      \begin{minipage}{.27\linewidth}
            \includegraphics[width=\linewidth]{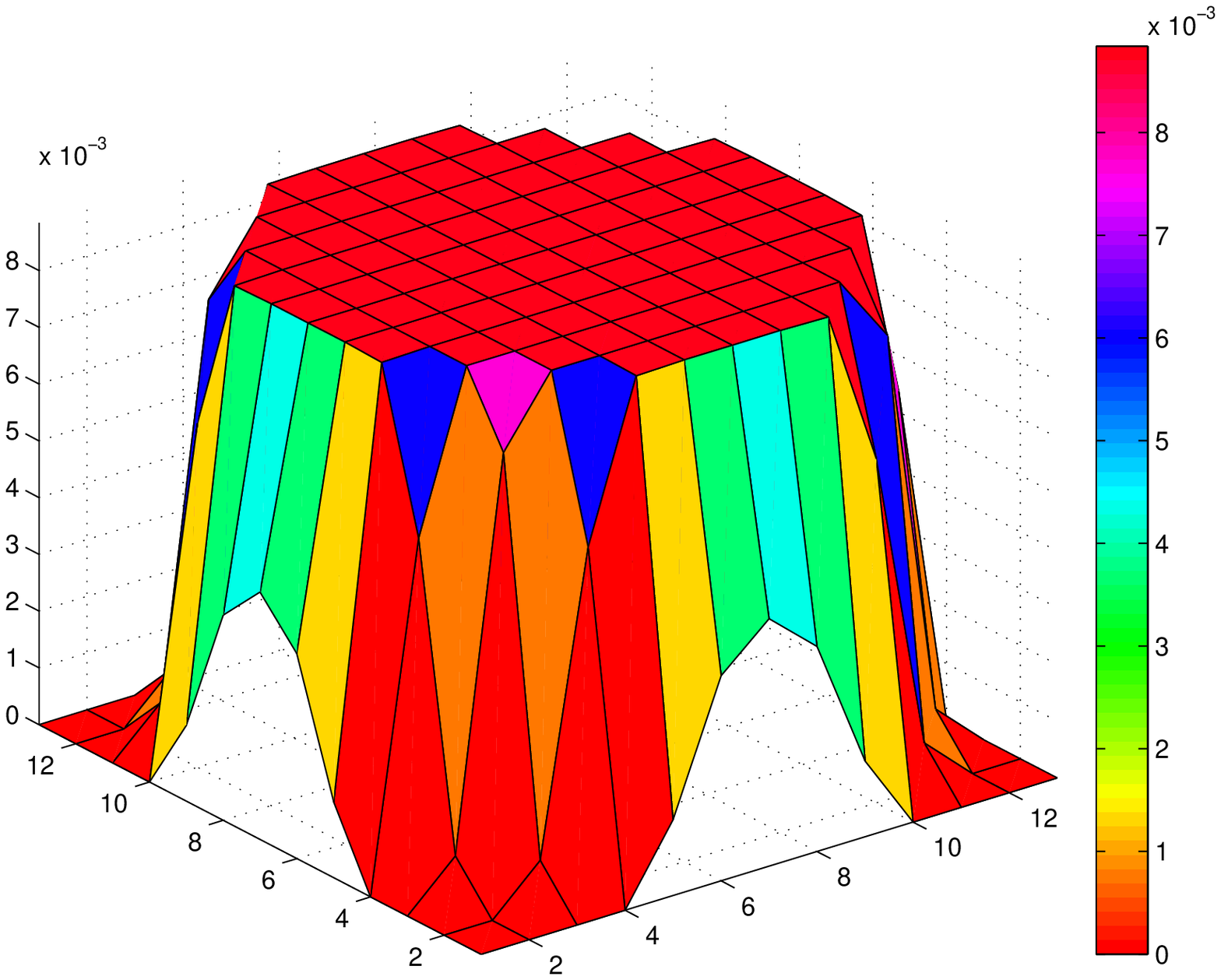}
      \end{minipage}
      \begin{minipage}{.27\linewidth}
            \includegraphics[width=\linewidth]{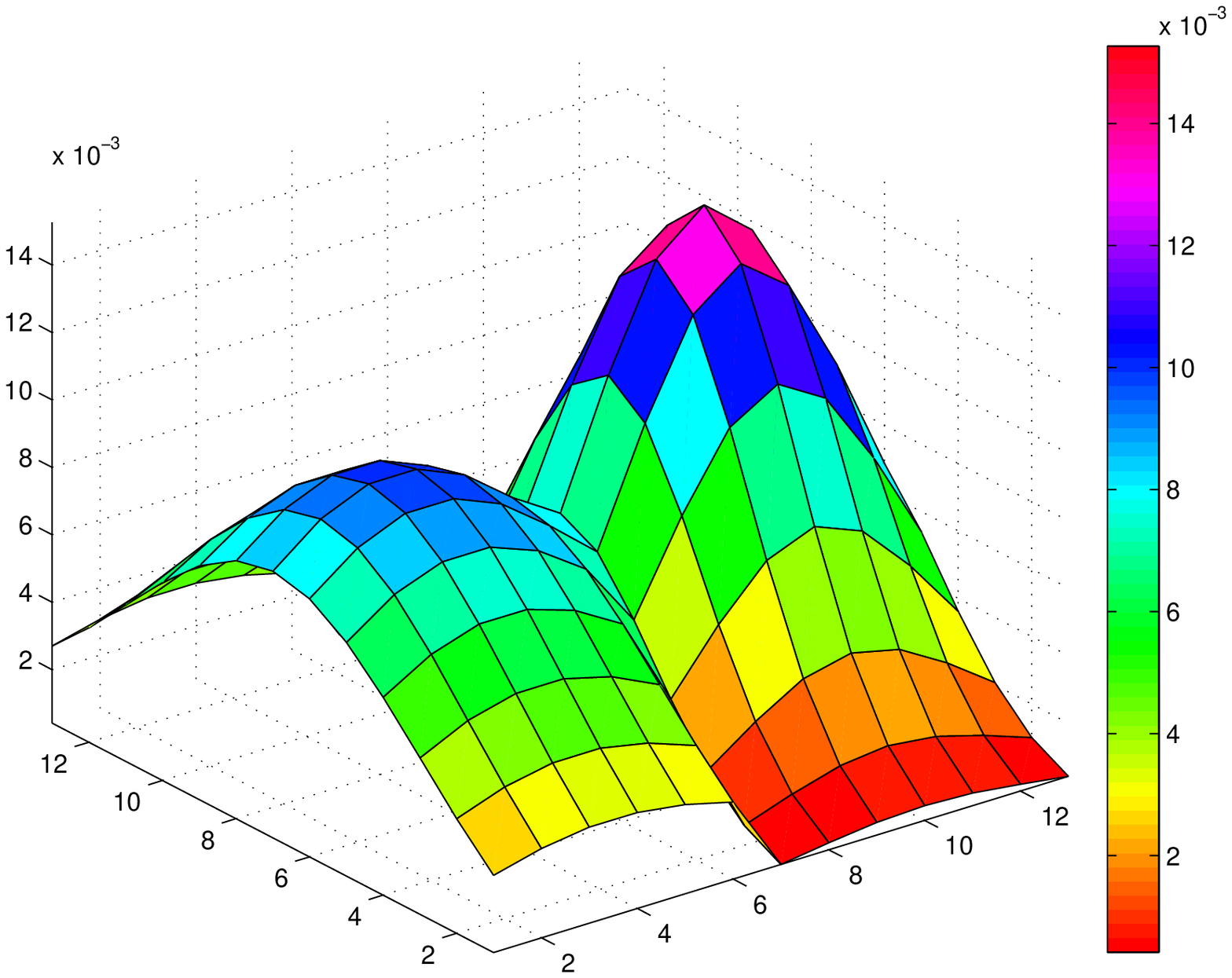}
      \end{minipage} \\
      \caption{Three simulated spatially-invariant PSFs ($\sharp1-\sharp3$) of size $13\times13$.}
      \label{blurkernels}
\end{figure}
\begin{figure}[!h]
      \centering
      \begin{minipage}{.31\linewidth}
            \includegraphics[width=\linewidth]{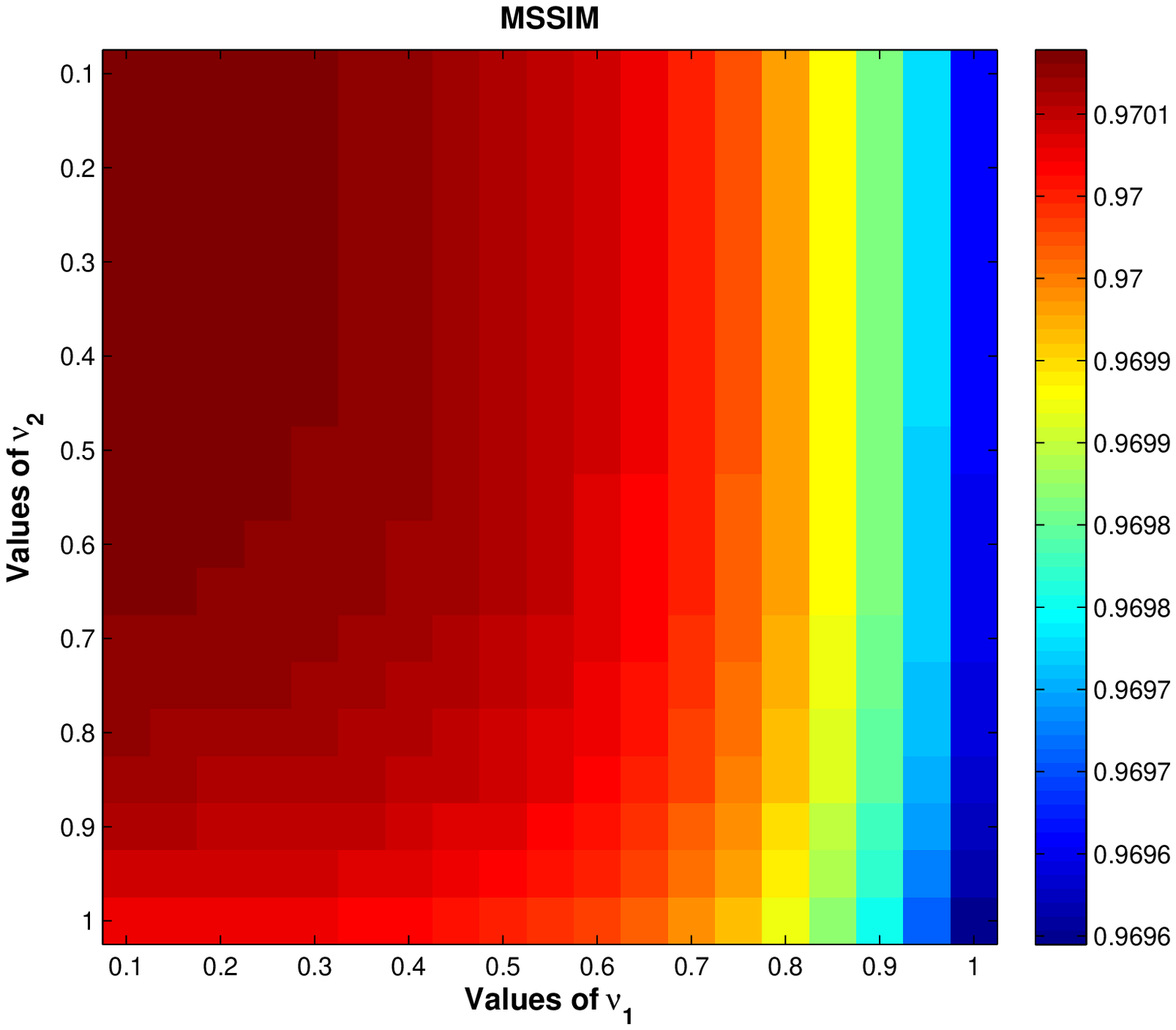}
      \end{minipage}
      \begin{minipage}{.31\linewidth}
            \includegraphics[width=\linewidth]{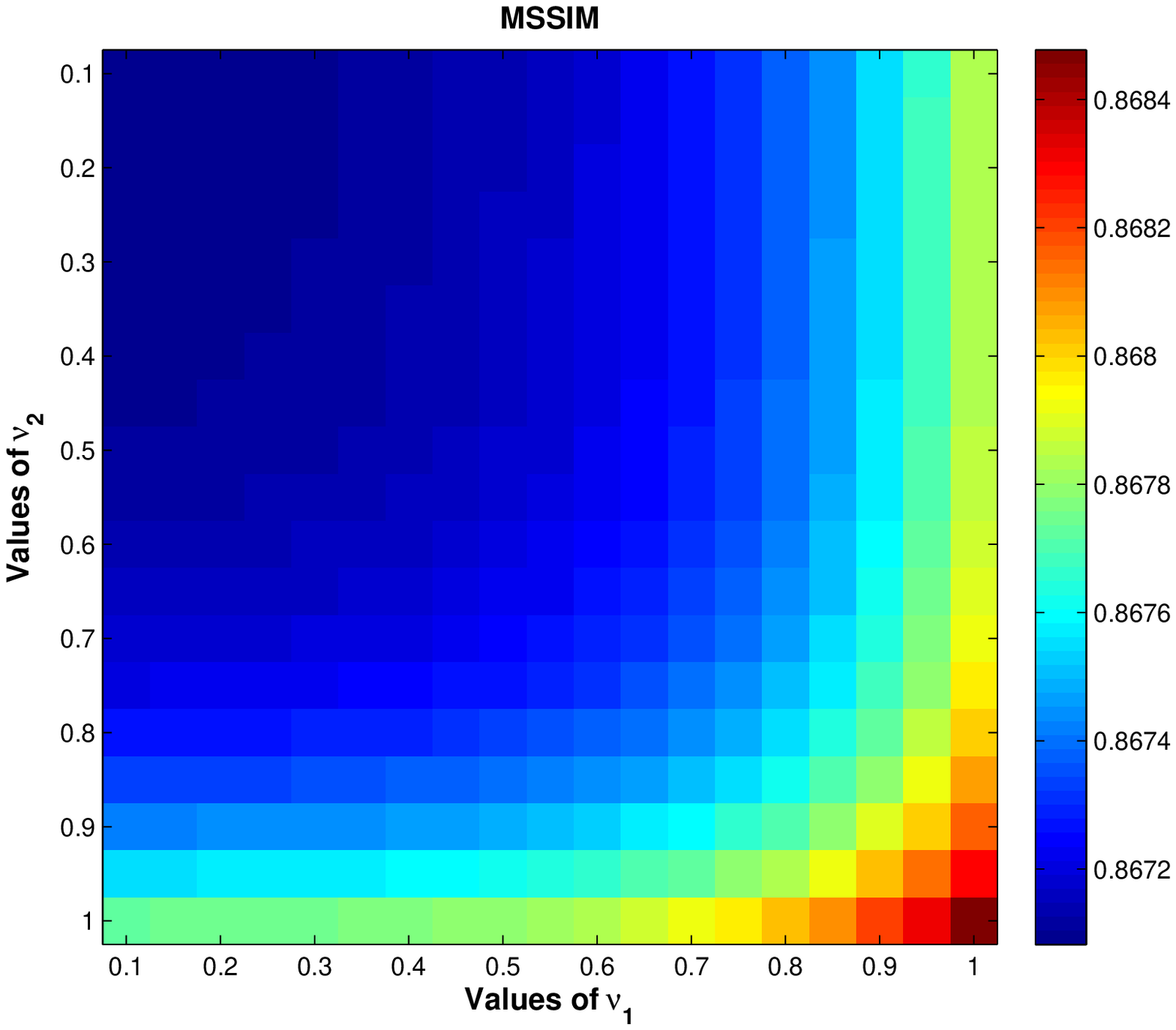}
      \end{minipage} \\
      \caption{Plot of the MSSIM values of deblurred images as functions of parameter pairs $\left(\nu_{1}, \nu_{2}\right)$. The original ``ADI" image in left experiment was only corrupted by PSF $\sharp 1$; whereas, in right experiment was simultaneously corrupted by PSF $\sharp 1$ and Gaussian noise with level $2\%$.}
      \label{PlotMSSIM}
\end{figure}
\begin{figure}[!h]
      \centering
      \begin{minipage}{.31\linewidth}
            \includegraphics[width=\linewidth]{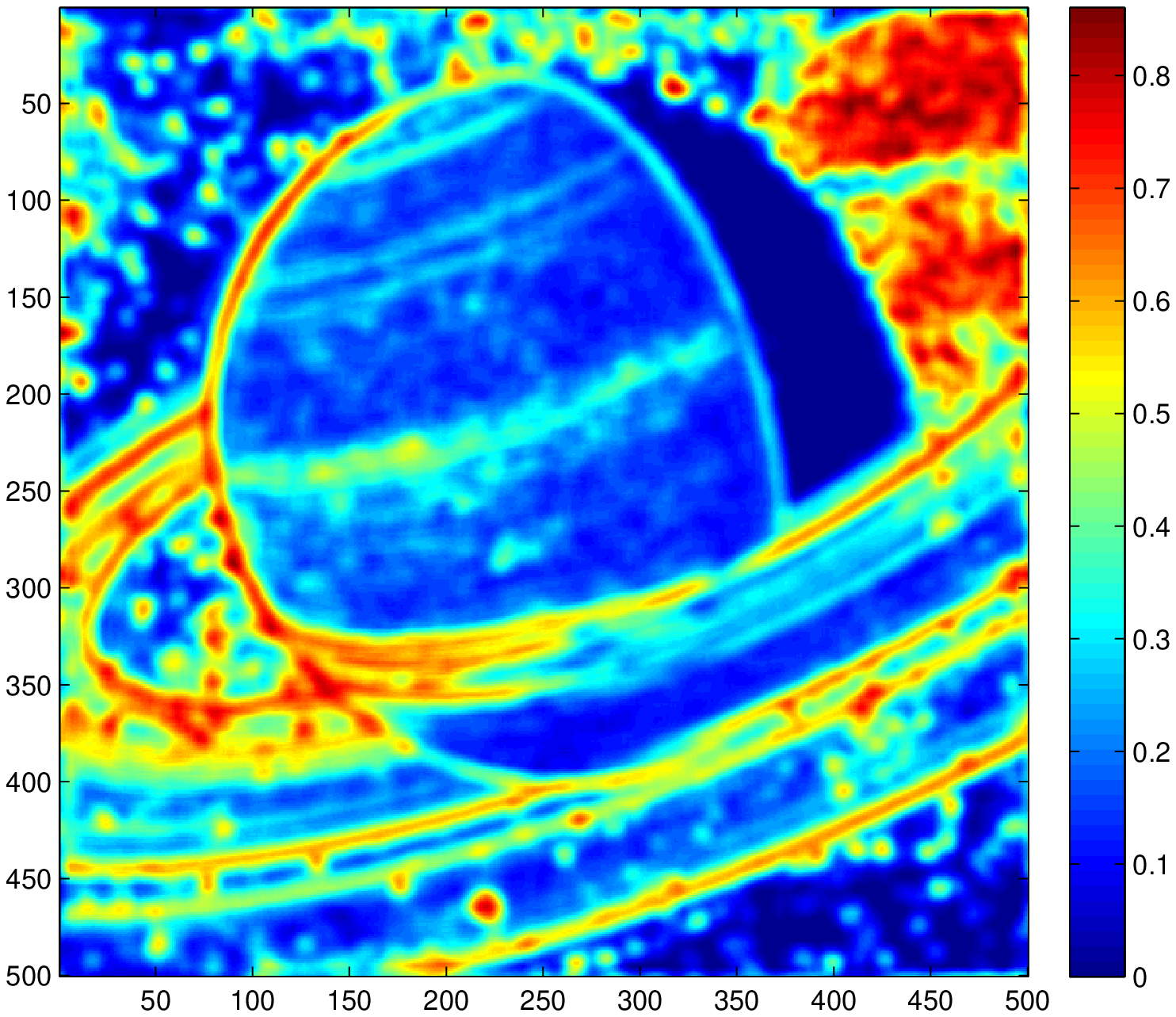}
      \end{minipage}
      \begin{minipage}{.31\linewidth}
            \includegraphics[width=\linewidth]{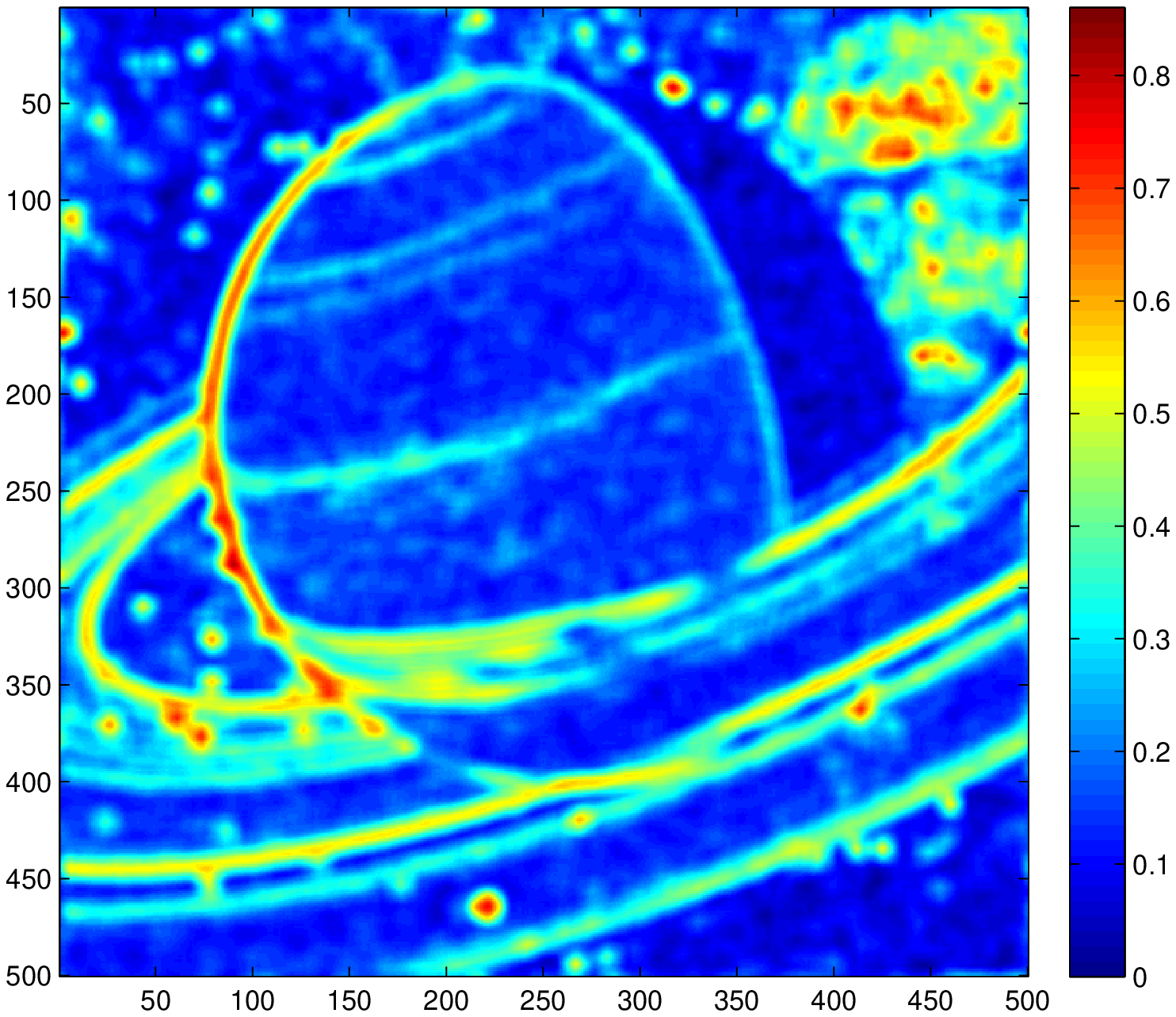}
      \end{minipage} \\
      \caption{Final results of the adaptive weighting function $\zeta$ correspond to the CNCHTV deblurring with $\nu_{1} = 0.55$ and $\nu_{2} = 0.55$ under different blurring conditions shown in Fig.\ref{PlotMSSIM}.}
      \label{PlotParameter}
\end{figure}
\section{Numerical Experiments and Analysis}
This section gives a detailed description of the qualitative and quantitative assessment of our proposed deblurring scheme. We select three images of size $500 \times 500$, known as fluorescence microscopy image (FMI), astronomical digital image (ADI) and remote sensing image (RSI), respectively. The deblurring results are compared with two state-of-the-art methods proposed by Krishnan et al. \cite{Krishnan} and Chan et al. \cite{ChanKhoshabeh}. The Mean Structural Similarity Index (MSSIM) {\cite{WangSSIM}} is used to measure deblurring performance. Fig.\ref{blurkernels} illustrates three simulated point spread functions (PSFs) used to generate synthetically degraded images. Fig.\ref{PlotMSSIM} plots the MSSIM values of deblurred images as functions of $\left(\nu_{1},\nu_{2}\right)$. In the case of blur degradation only, the pair $(0.1,0.1)$ generates the best restoration performance. In contrast, the pair $(1,1)$ is the optimal selection under the existence of PSF and Gaussian noise. To maintain a successful balance, an experiential choice $(0.55,0.55)$ is used throughout the rest of this paper. The values of corresponding $\zeta$ are presented in Fig.\ref{PlotParameter}. It is obvious that the adaptive $\zeta$ obtained by (\ref{weighting}) could exactly detect texture and homogeneous regions to yield good deblurring results.
\begin{figure}[H]
\centering
      \begin{minipage}{.185\linewidth}
            \includegraphics[width=\linewidth]{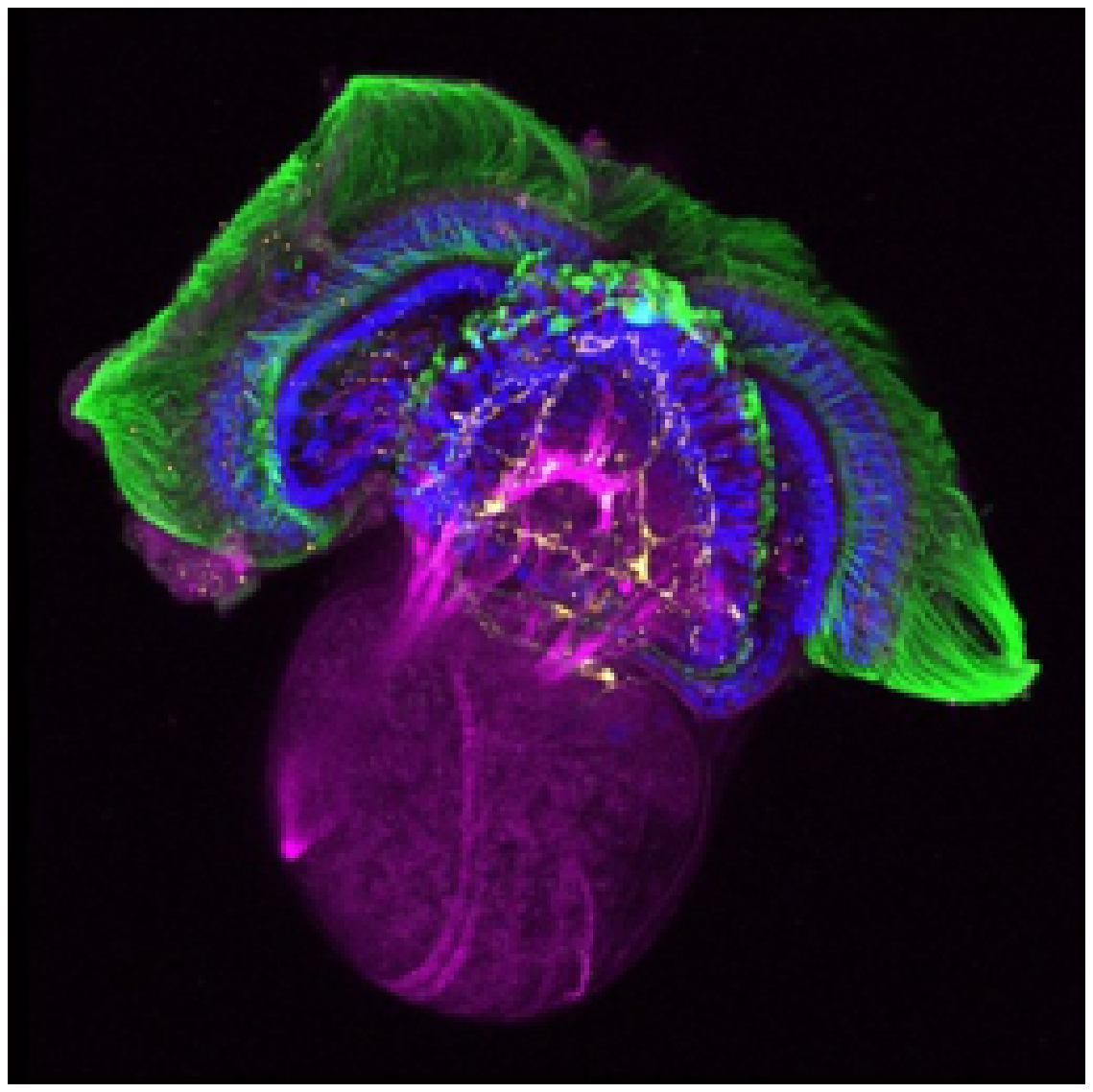}
      \end{minipage}
      \begin{minipage}{.185\linewidth}
            \includegraphics[width=\linewidth]{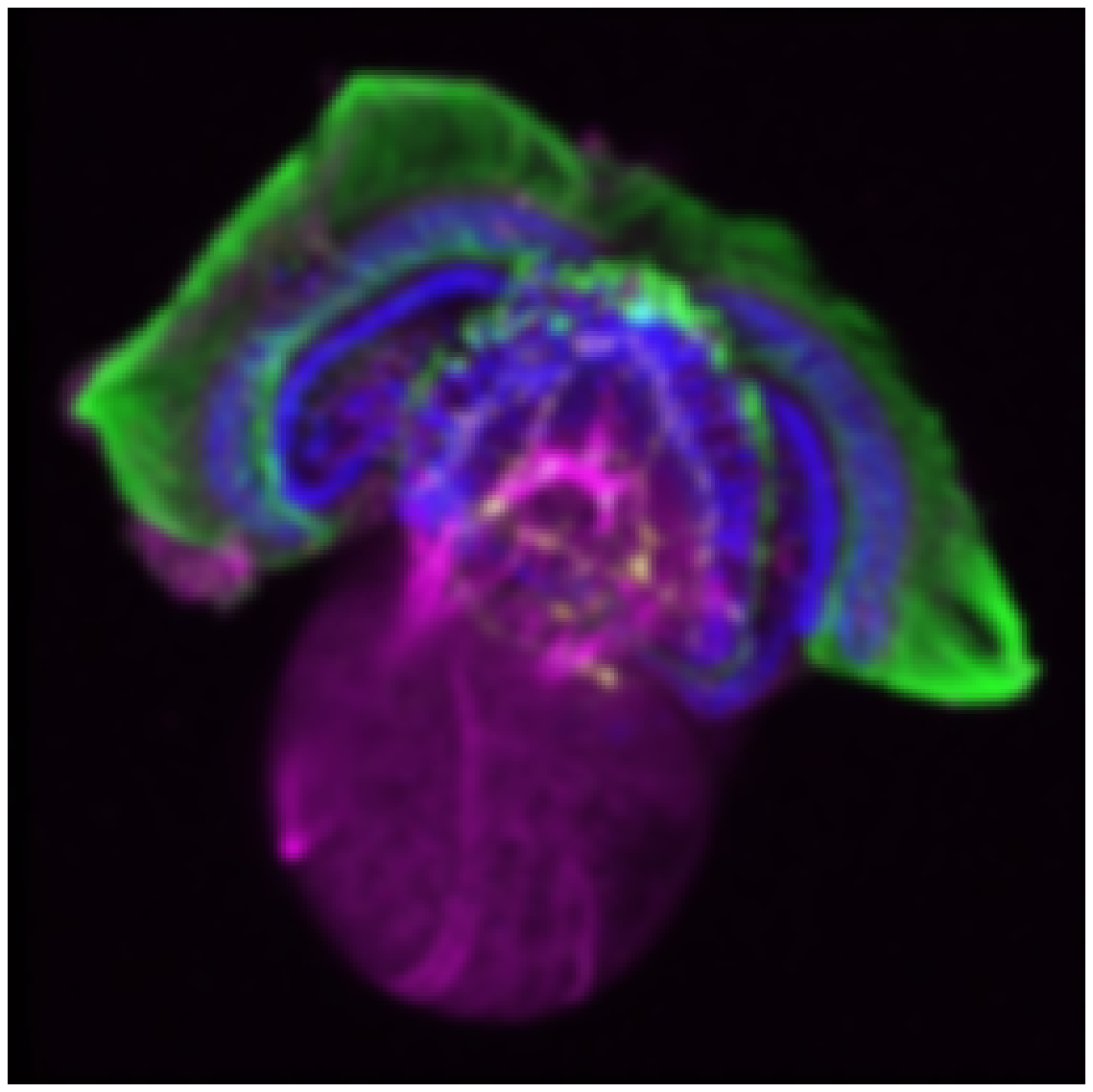}
      \end{minipage}
       \begin{minipage}{.185\linewidth}
            \includegraphics[width=\linewidth]{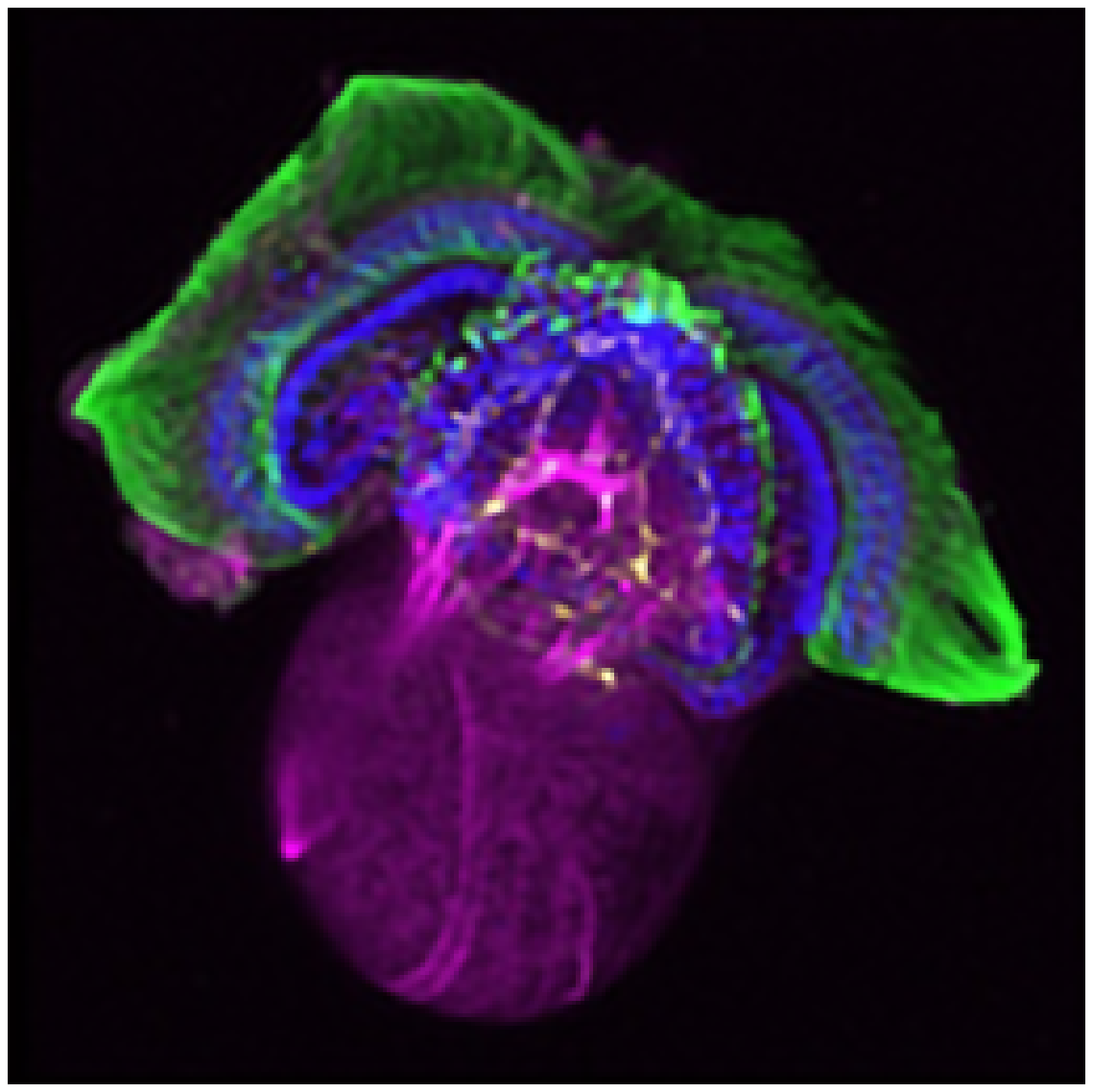}
      \end{minipage}
      \begin{minipage}{.185\linewidth}
            \includegraphics[width=\linewidth]{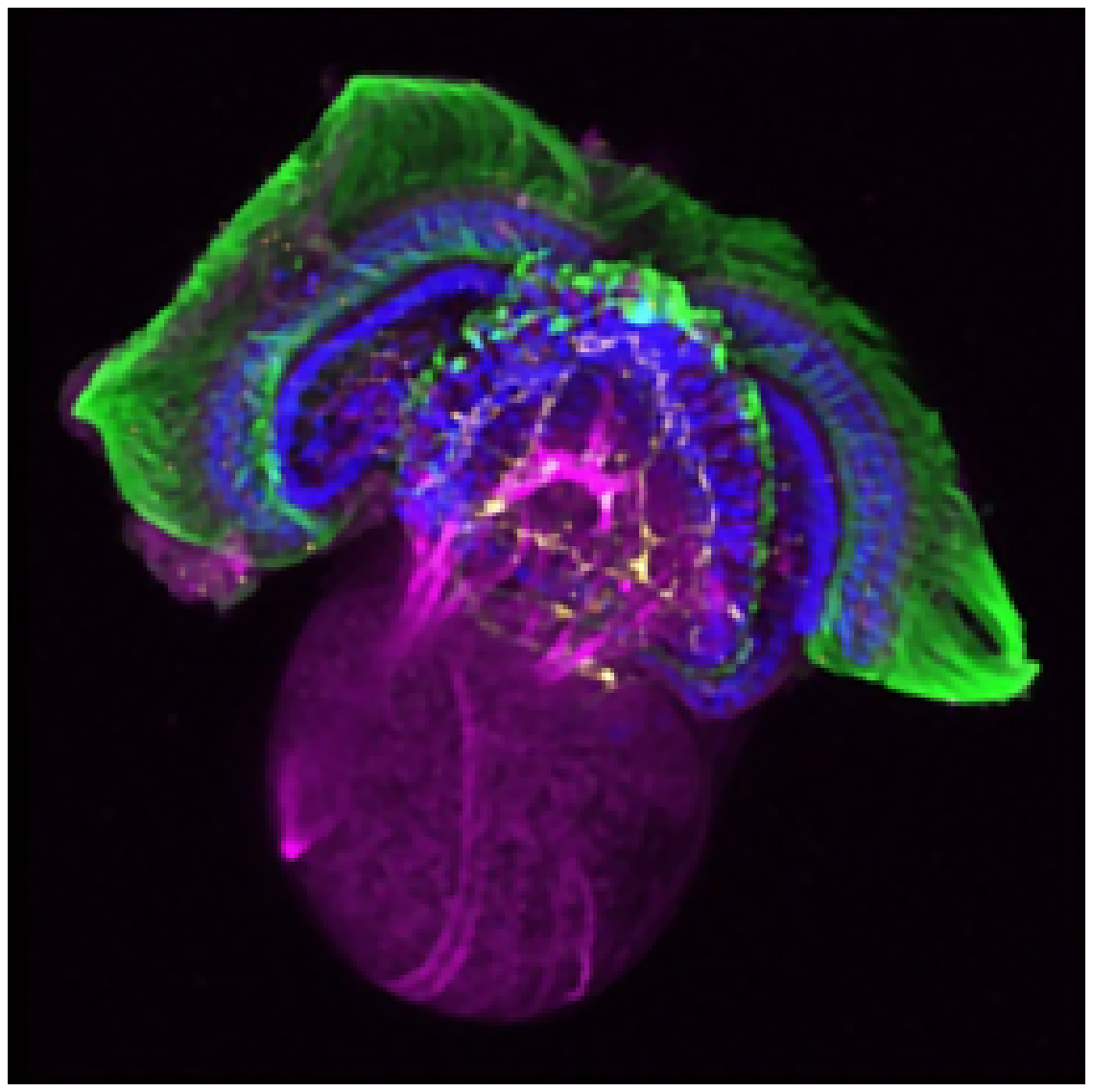}
      \end{minipage}
      \begin{minipage}{.185\linewidth}
            \includegraphics[width=\linewidth]{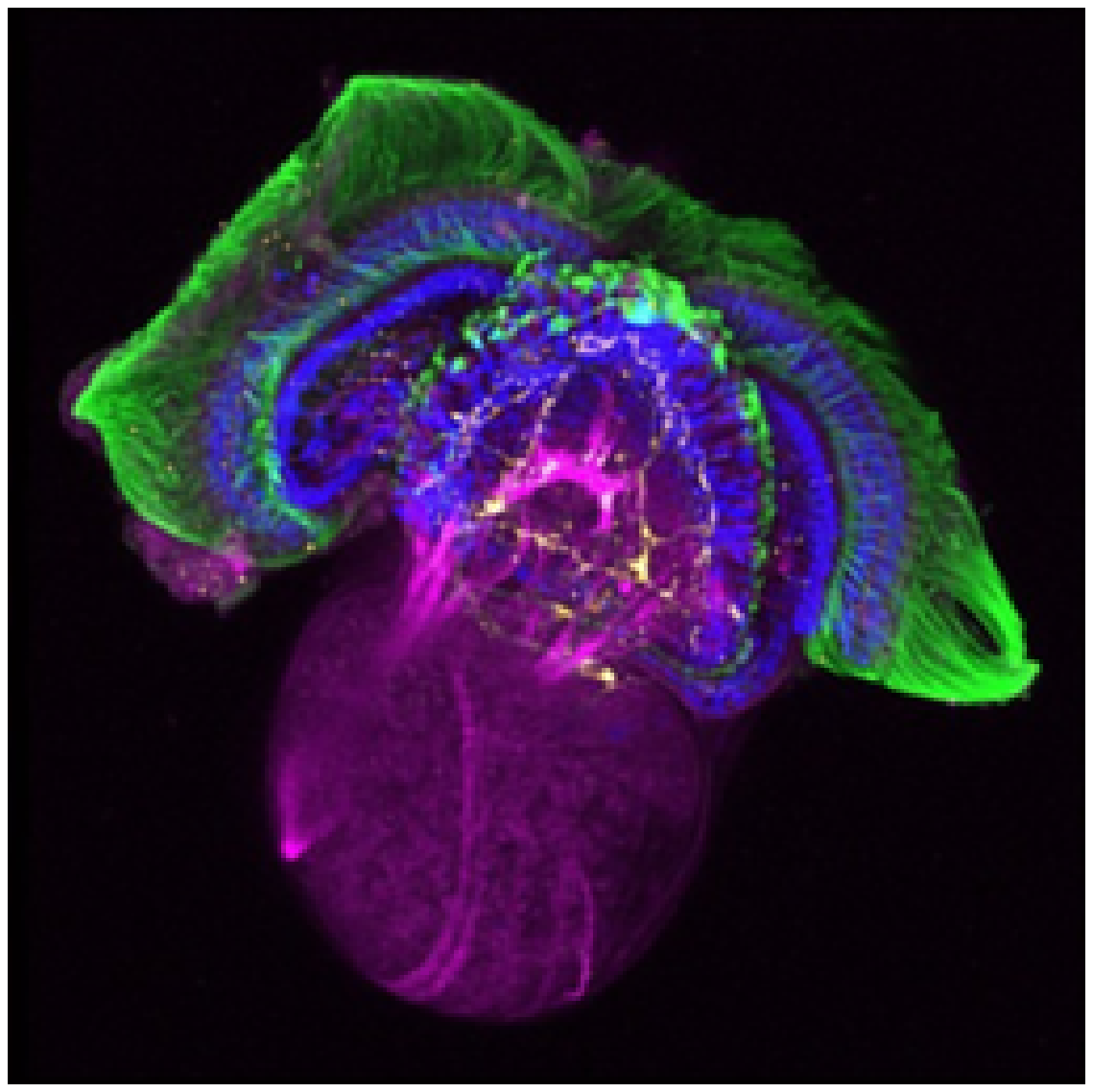}
      \end{minipage}\\
      \begin{minipage}{.185\linewidth}
            \includegraphics[width=\linewidth]{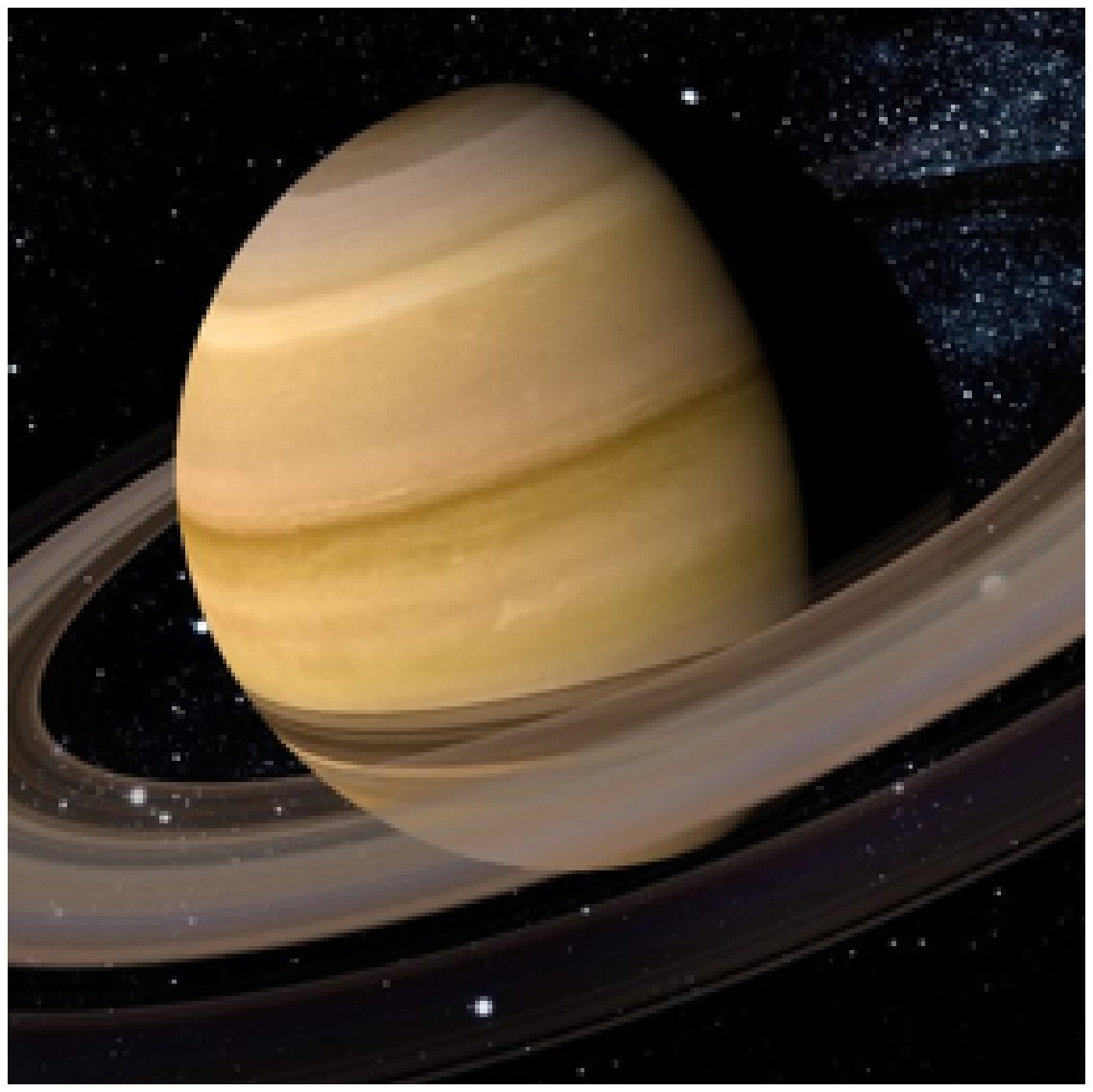}
      \end{minipage}
      \begin{minipage}{.185\linewidth}
            \includegraphics[width=\linewidth]{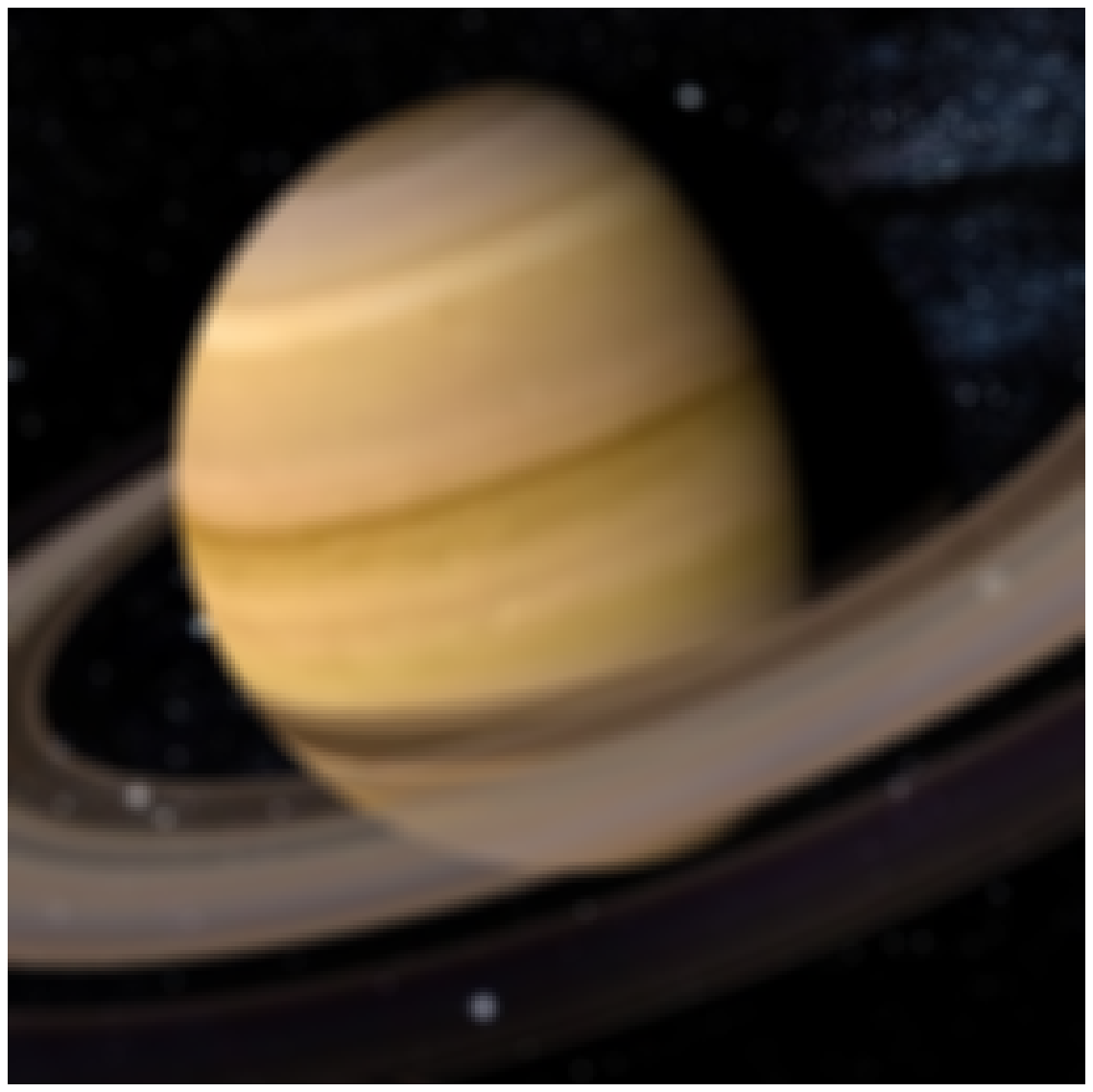}
      \end{minipage}
      \begin{minipage}{.185\linewidth}
            \includegraphics[width=\linewidth]{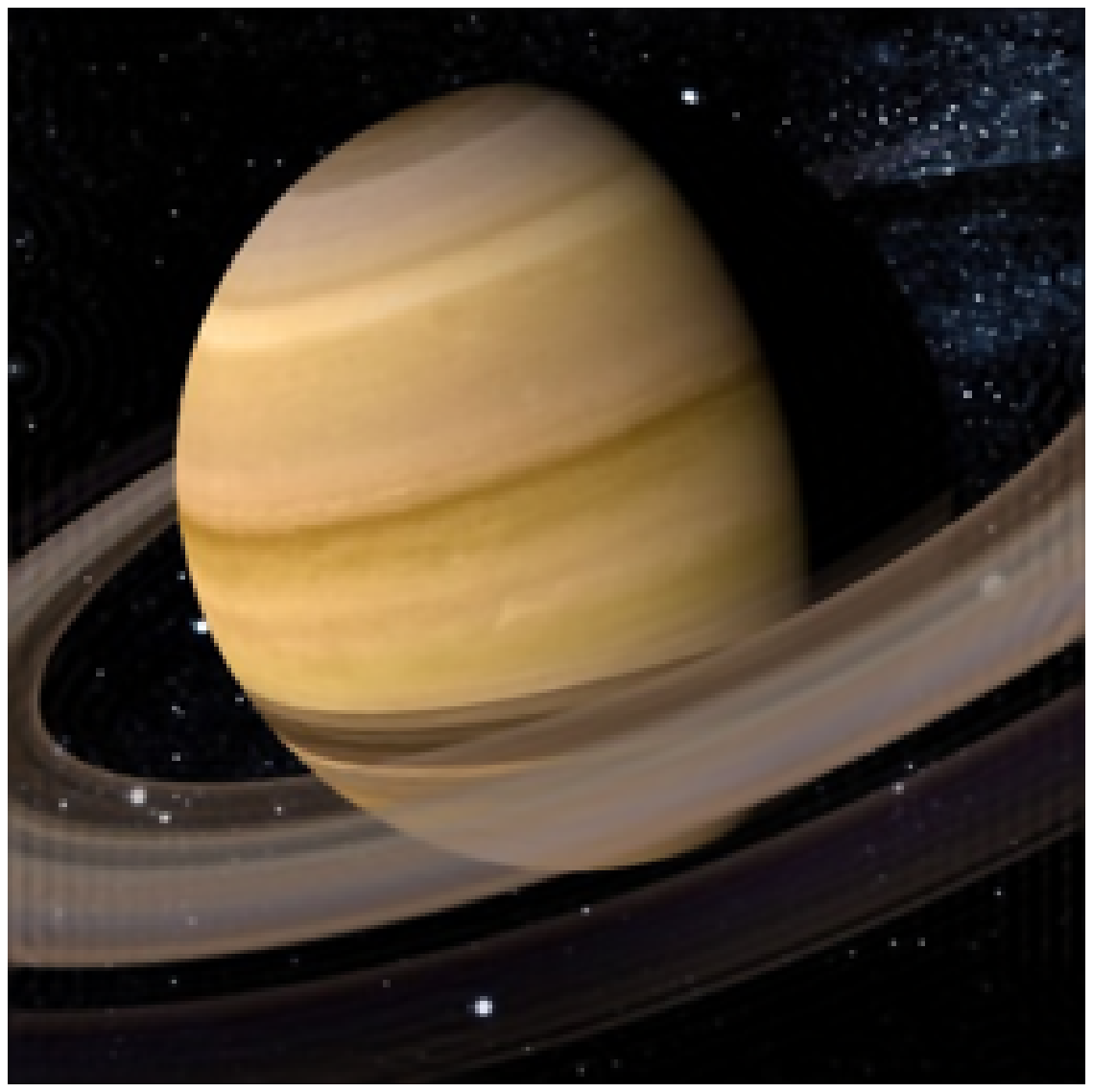}
      \end{minipage}
      \begin{minipage}{.185\linewidth}
            \includegraphics[width=\linewidth]{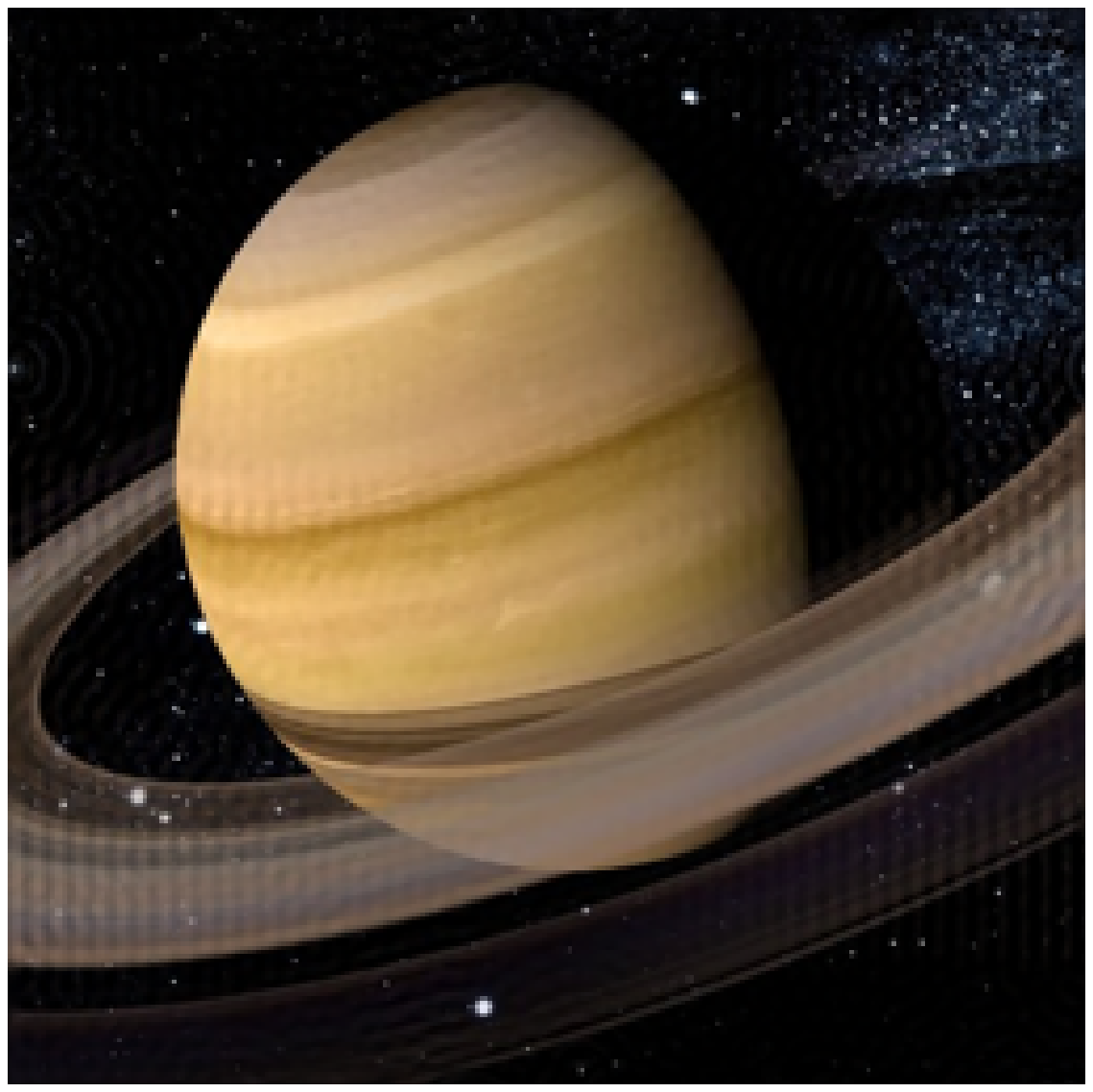}
      \end{minipage}
      \begin{minipage}{.185\linewidth}
            \includegraphics[width=\linewidth]{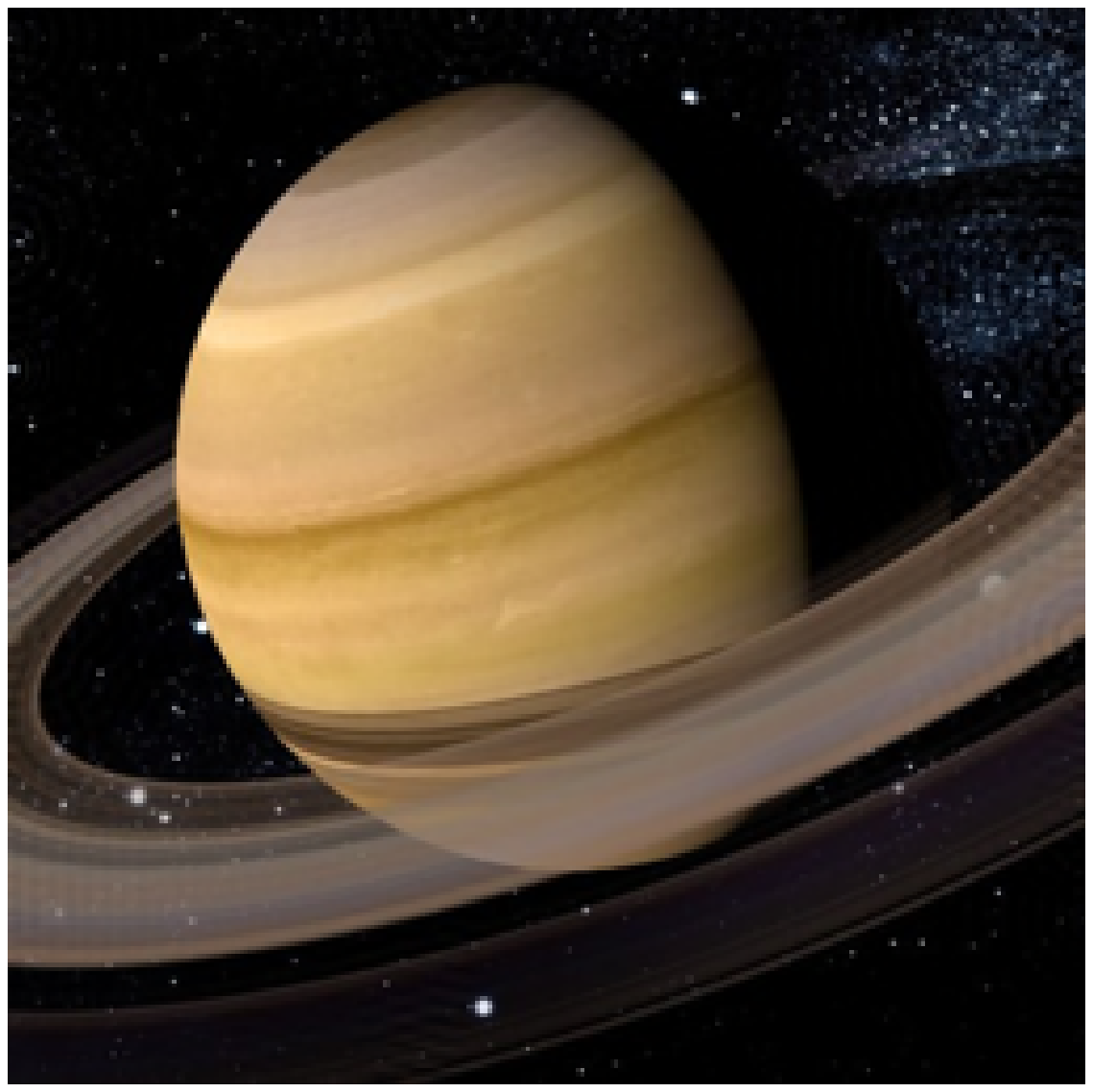}
      \end{minipage}\\
      \begin{minipage}{.185\linewidth}
            \includegraphics[width=\linewidth]{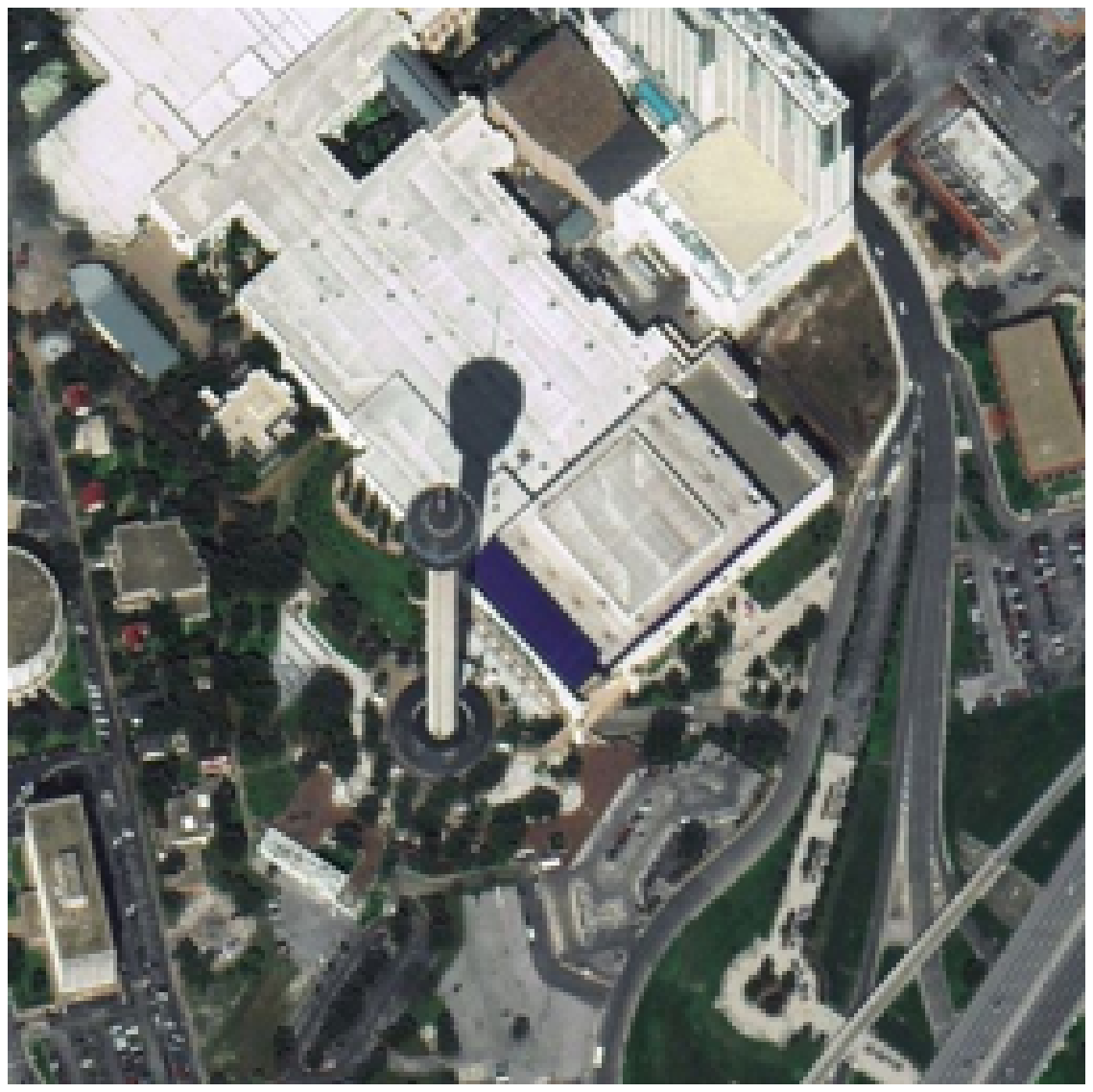}
            \centerline{{(a)}}
      \end{minipage}
      \begin{minipage}{.185\linewidth}
            \includegraphics[width=\linewidth]{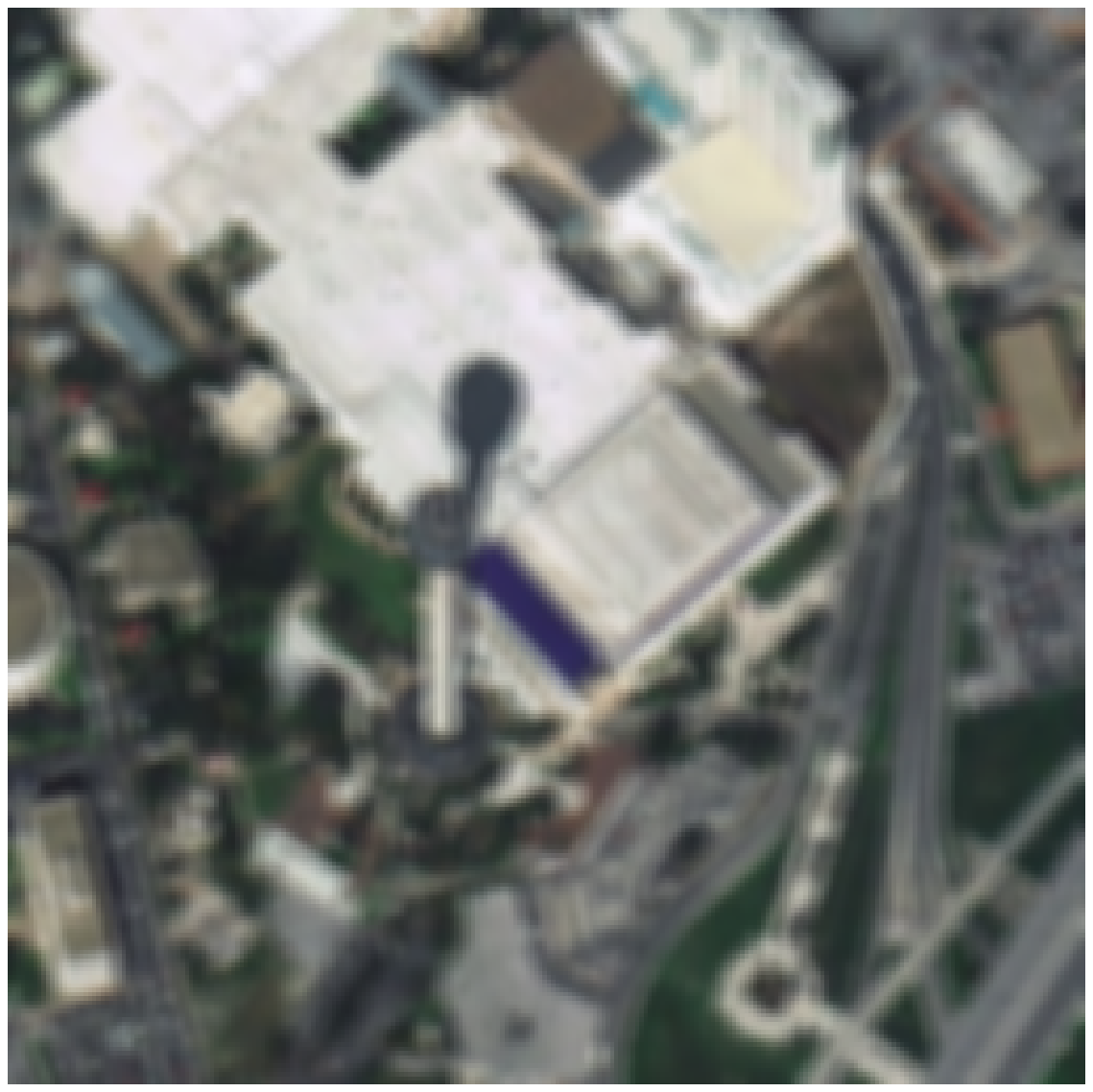}
            \centerline{{(b)}}
      \end{minipage}
      \begin{minipage}{.185\linewidth}
            \includegraphics[width=\linewidth]{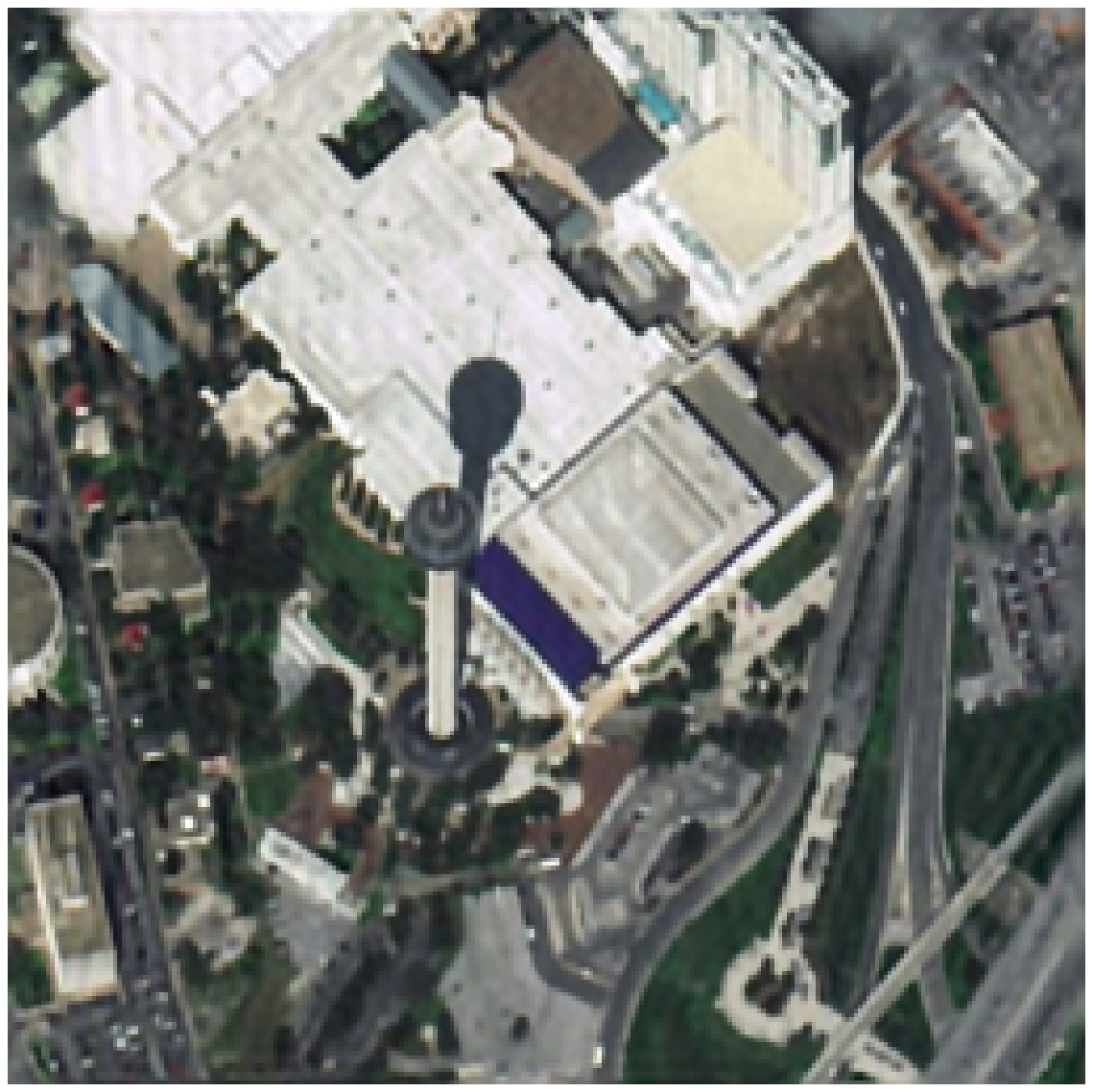}
            \centerline{{(c)}}
      \end{minipage}
      \begin{minipage}{.185\linewidth}
            \includegraphics[width=\linewidth]{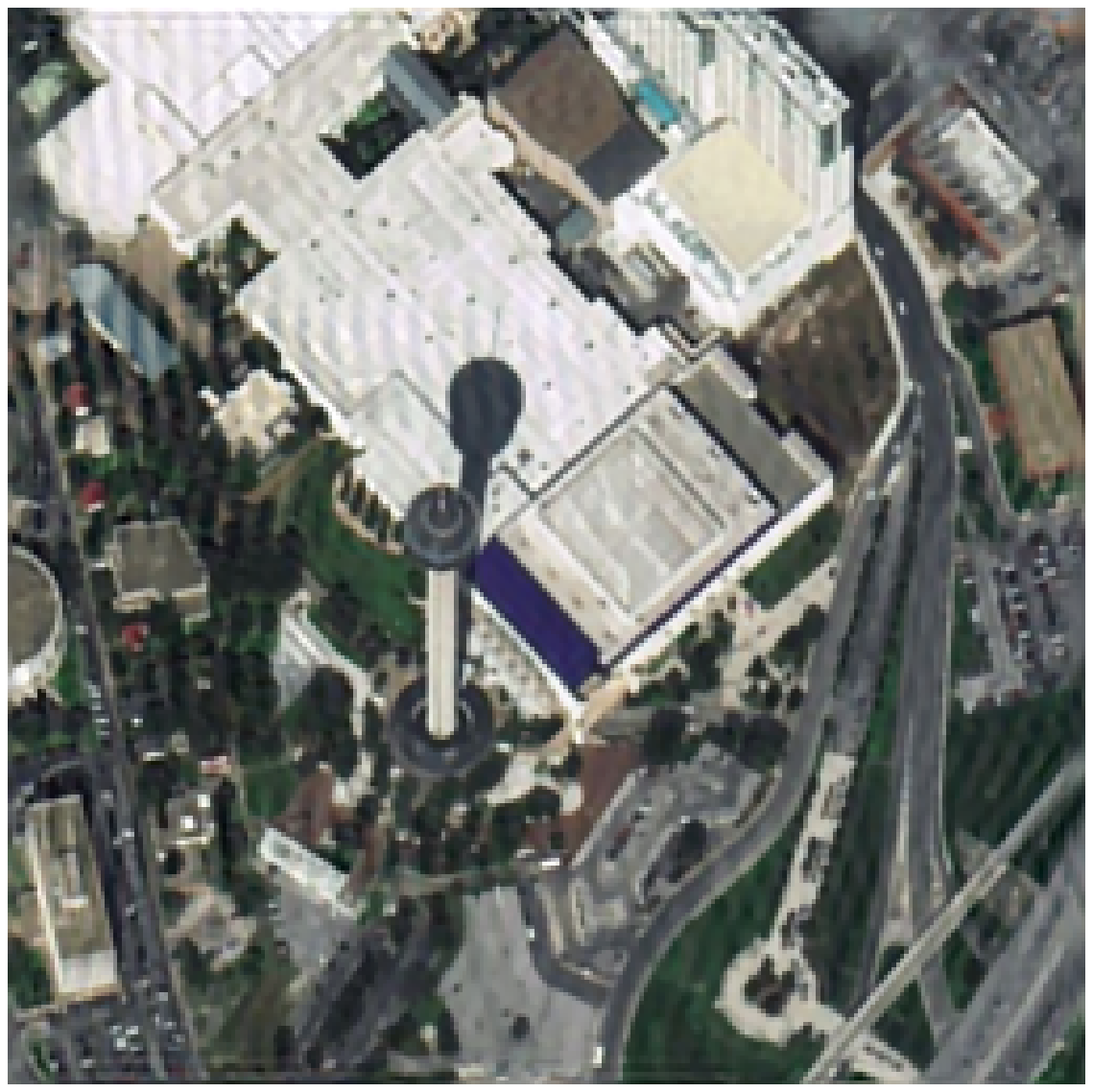}
            \centerline{{(d)}}
      \end{minipage}
      \begin{minipage}{.185\linewidth}
            \includegraphics[width=\linewidth]{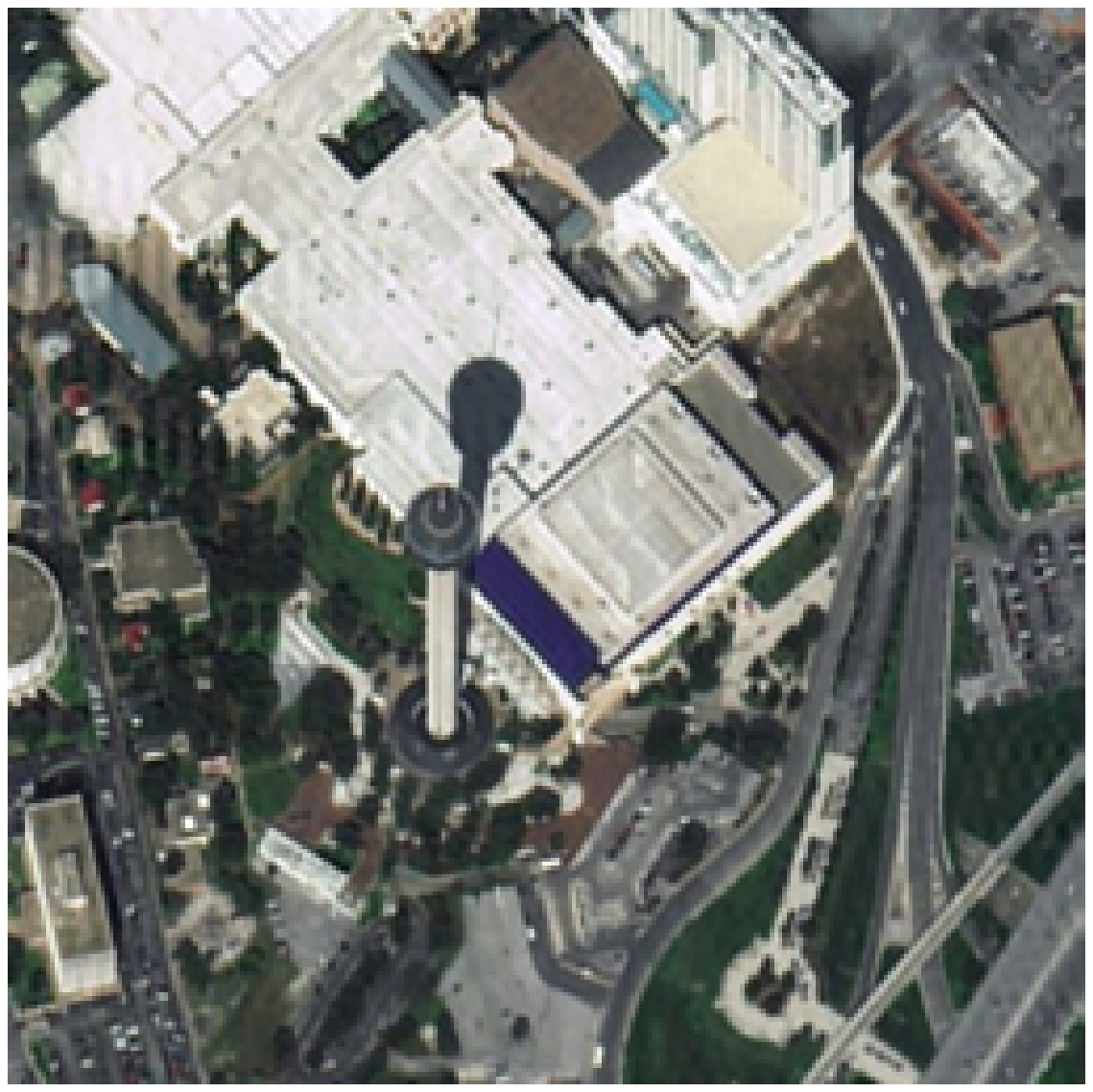}
            \centerline{{(e)}}
      \end{minipage}\\
      \caption{Comparison of deblurring results using different methods on ``FMI", ``ADI" and ``RSI" images (from top to bottom). These three images are only degraded by spatially-invariant PSFs $\sharp 1$, $\sharp 2$ and $\sharp 3$, respectively. From left to right, the (a) original image, (b) blurred image, deblurred versions of (c) Krishnan's method \cite{Krishnan}, (d) Chan's method \cite{ChanKhoshabeh} and (e) our CNCHTV model are displayed (The images are best viewed in full-screen mode).}
      \label{noisefree}
\end{figure}
%
%
%
\subsection{Noise-free Image Deblurring}
In the case of noise-free condition, the original images and their blurred/deblurred versions are shown in Fig.\ref{noisefree}. The parameters $\mu = 5 \times 10^{5}$ and $\beta_{i} = 1 \times 10^{2} ( i=1,2,3 )$ are selected empirically. The experimental results show that Krishnan's method \cite{Krishnan} with $\nu_{1} = \frac{1}{2}$ can suppress ringing artifacts effectively, but easily results in over-smoothing of fine details. In contrast, ringing artifacts generated by Chan's method \cite{ChanKhoshabeh} can lead to perceptible degradation of image quality. As shown in Fig.\ref{noisefree}(e), our proposed deblurring scheme significantly improves the visual deblurring quality. Thus the CNCHTV can keep a good balance between preserving image details and alleviating ringing artifacts.
\subsection{Image Deblurring with Gaussian Noise}

\begin{figure}
\centering
      \begin{minipage}{.185\linewidth}
            \includegraphics[width=\linewidth]{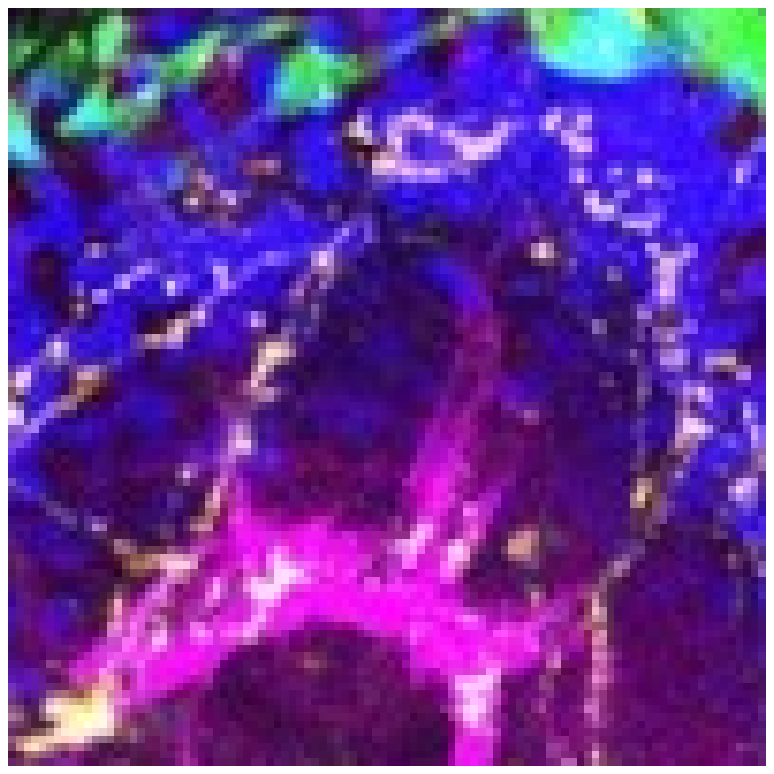}
      \end{minipage}
      \begin{minipage}{.185\linewidth}
            \includegraphics[width=\linewidth]{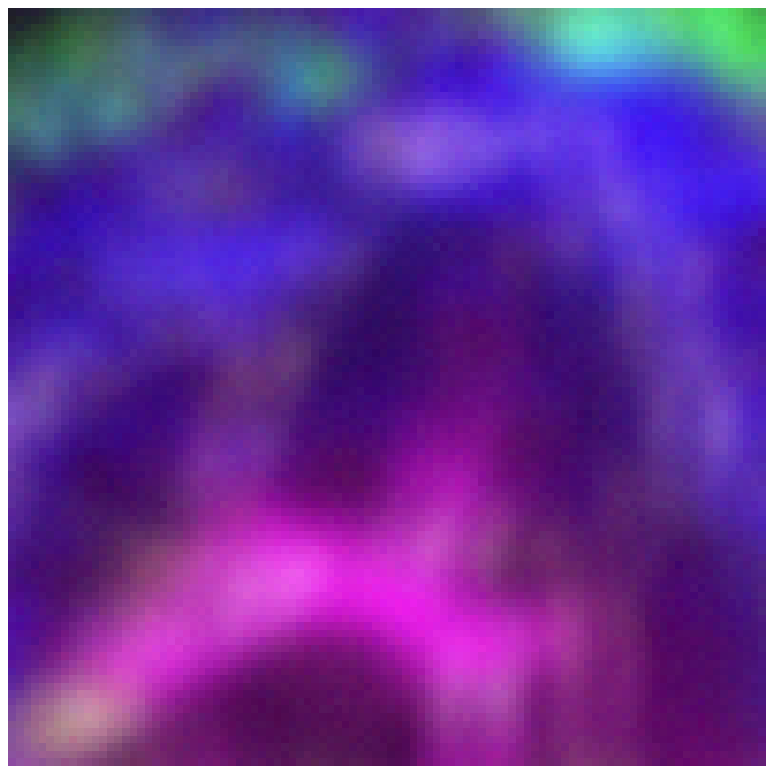}
      \end{minipage}
      \begin{minipage}{.185\linewidth}
            \includegraphics[width=\linewidth]{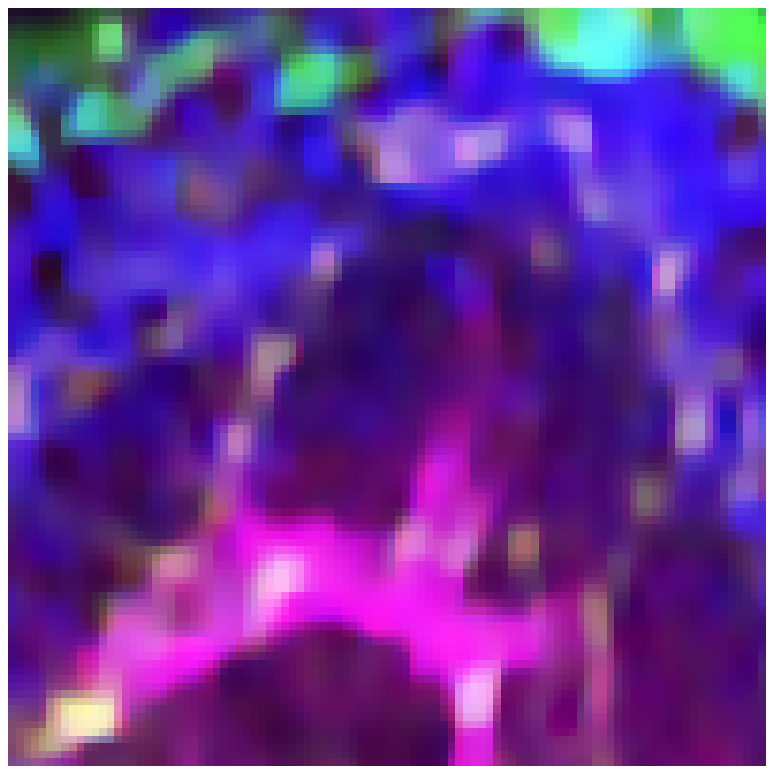}
      \end{minipage}
      \begin{minipage}{.185\linewidth}
            \includegraphics[width=\linewidth]{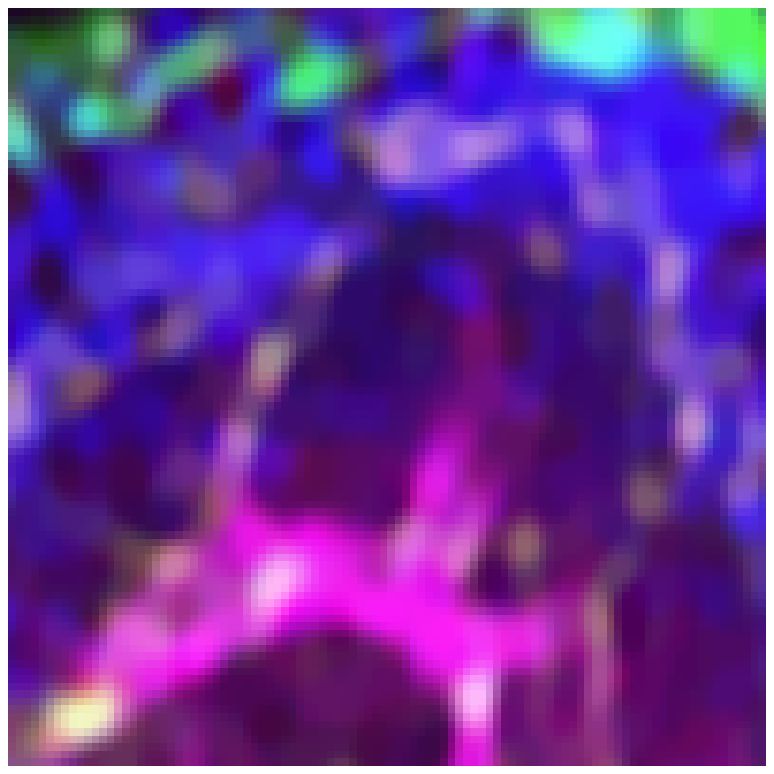}
      \end{minipage}
      \begin{minipage}{.185\linewidth}
            \includegraphics[width=\linewidth]{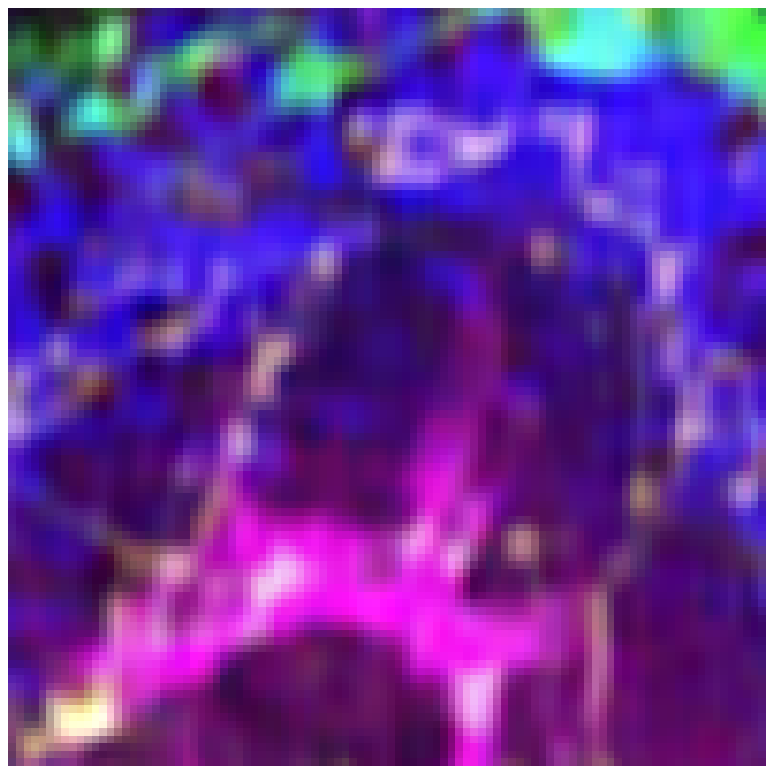}
      \end{minipage}\\
            \begin{minipage}{.185\linewidth}
            \includegraphics[width=\linewidth]{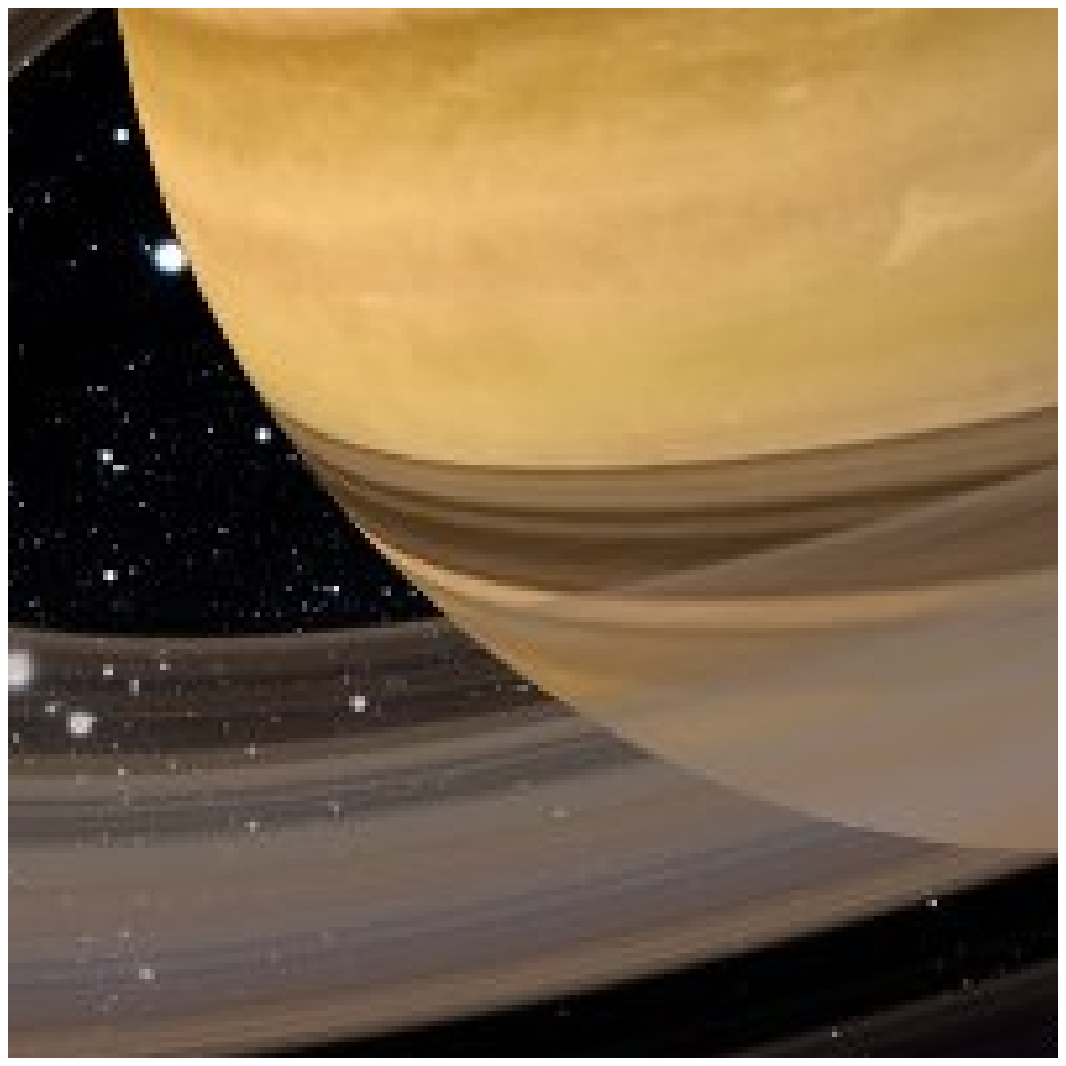}
      \end{minipage}
      \begin{minipage}{.185\linewidth}
            \includegraphics[width=\linewidth]{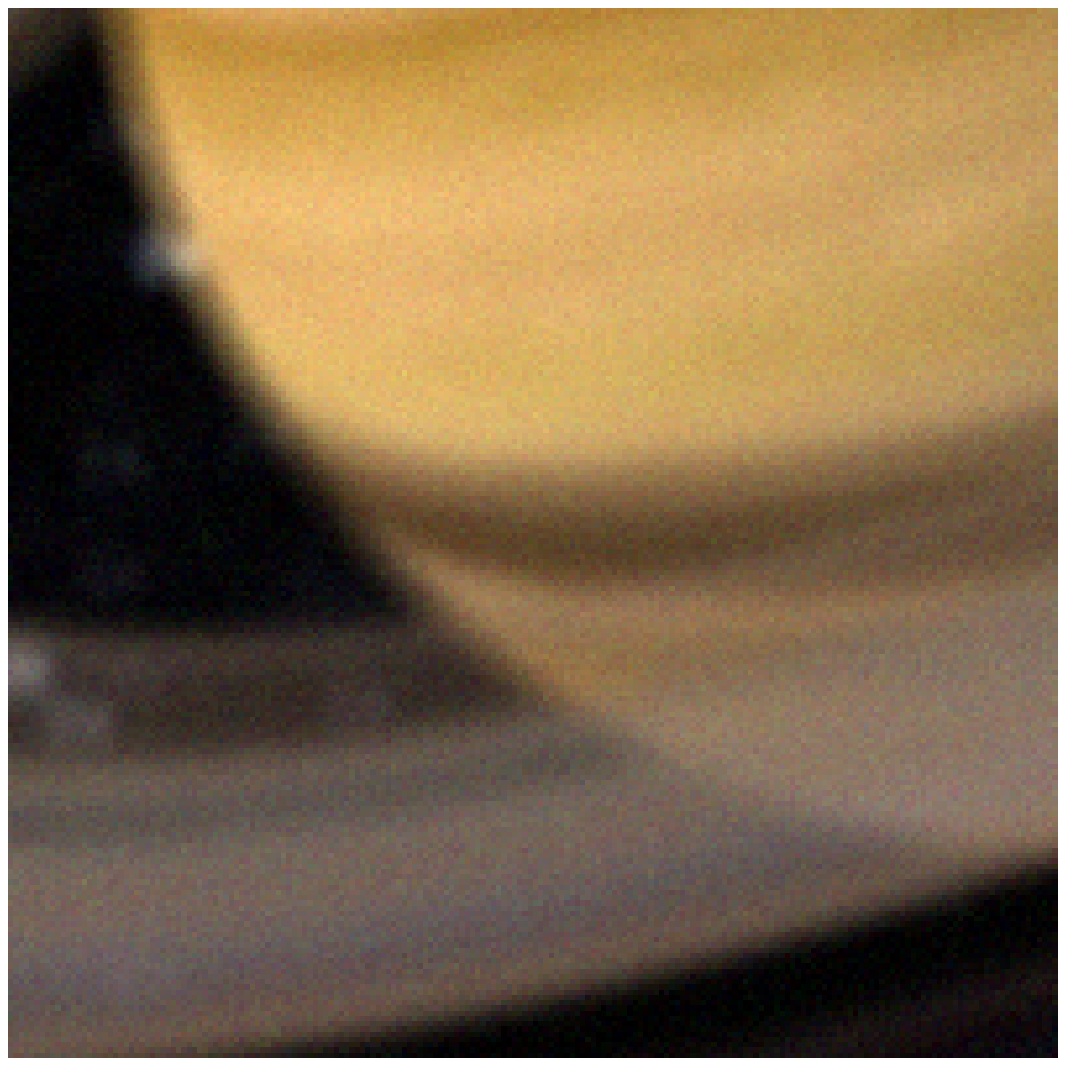}
      \end{minipage}
      \begin{minipage}{.185\linewidth}
            \includegraphics[width=\linewidth]{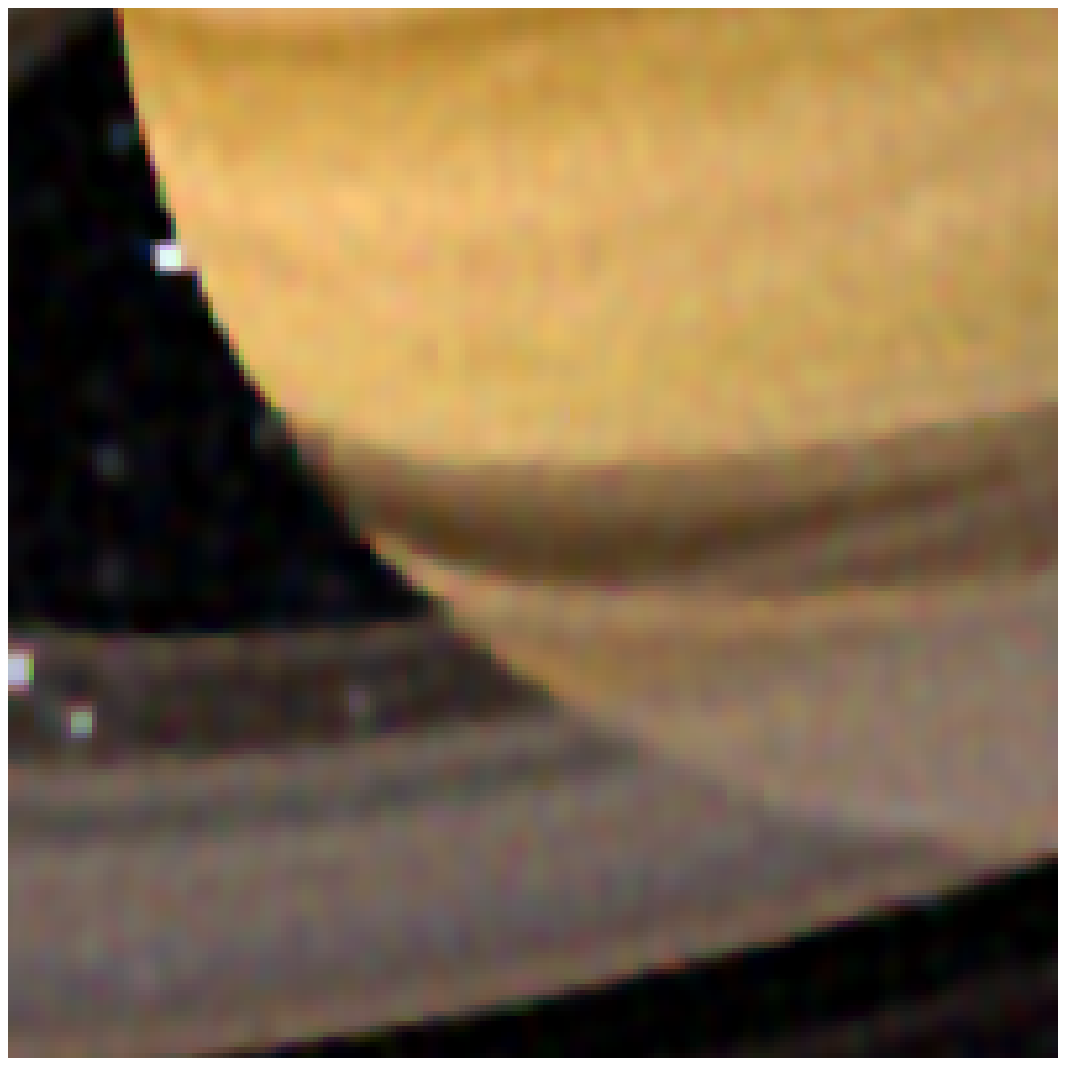}
      \end{minipage}
      \begin{minipage}{.185\linewidth}
            \includegraphics[width=\linewidth]{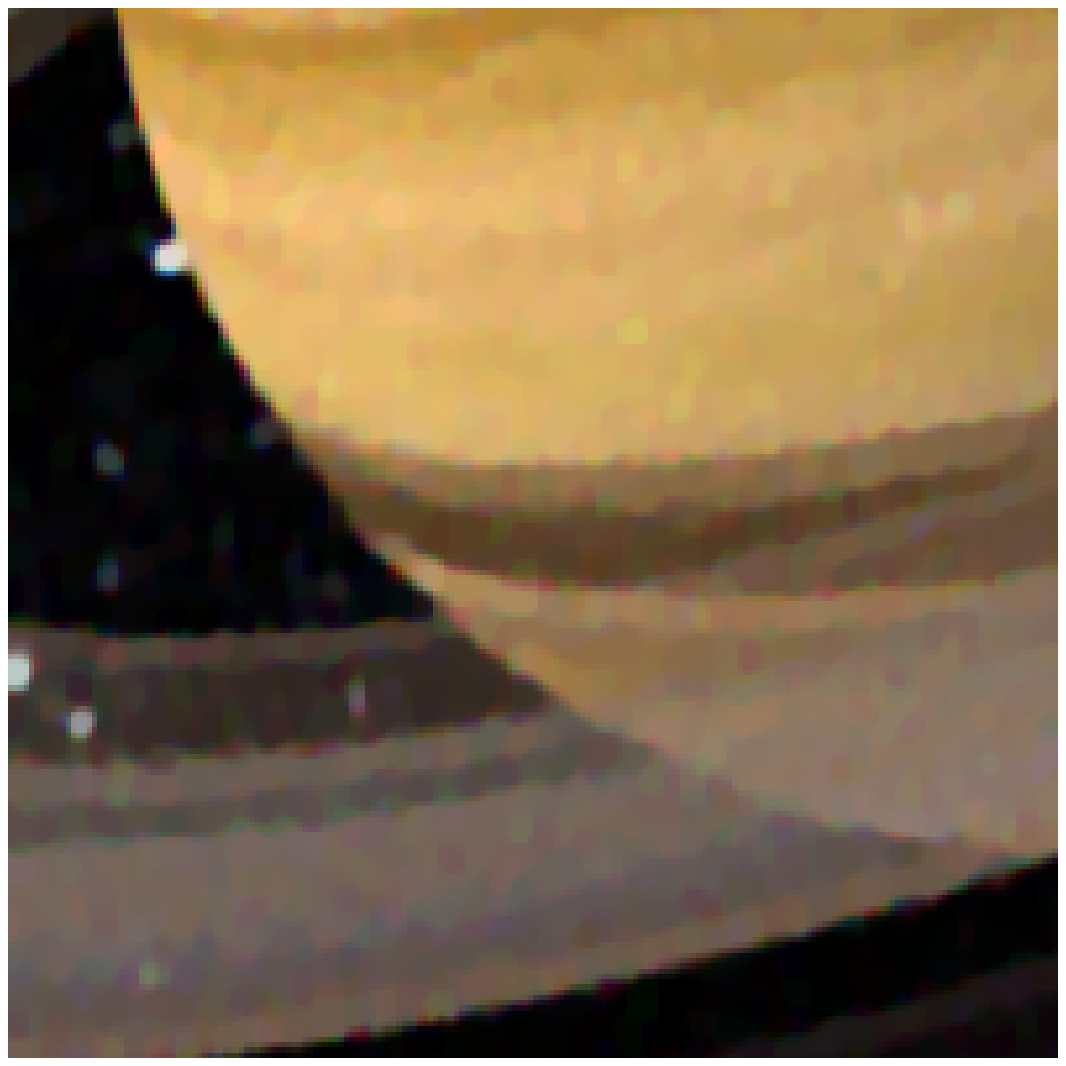}
      \end{minipage}
      \begin{minipage}{.185\linewidth}
            \includegraphics[width=\linewidth]{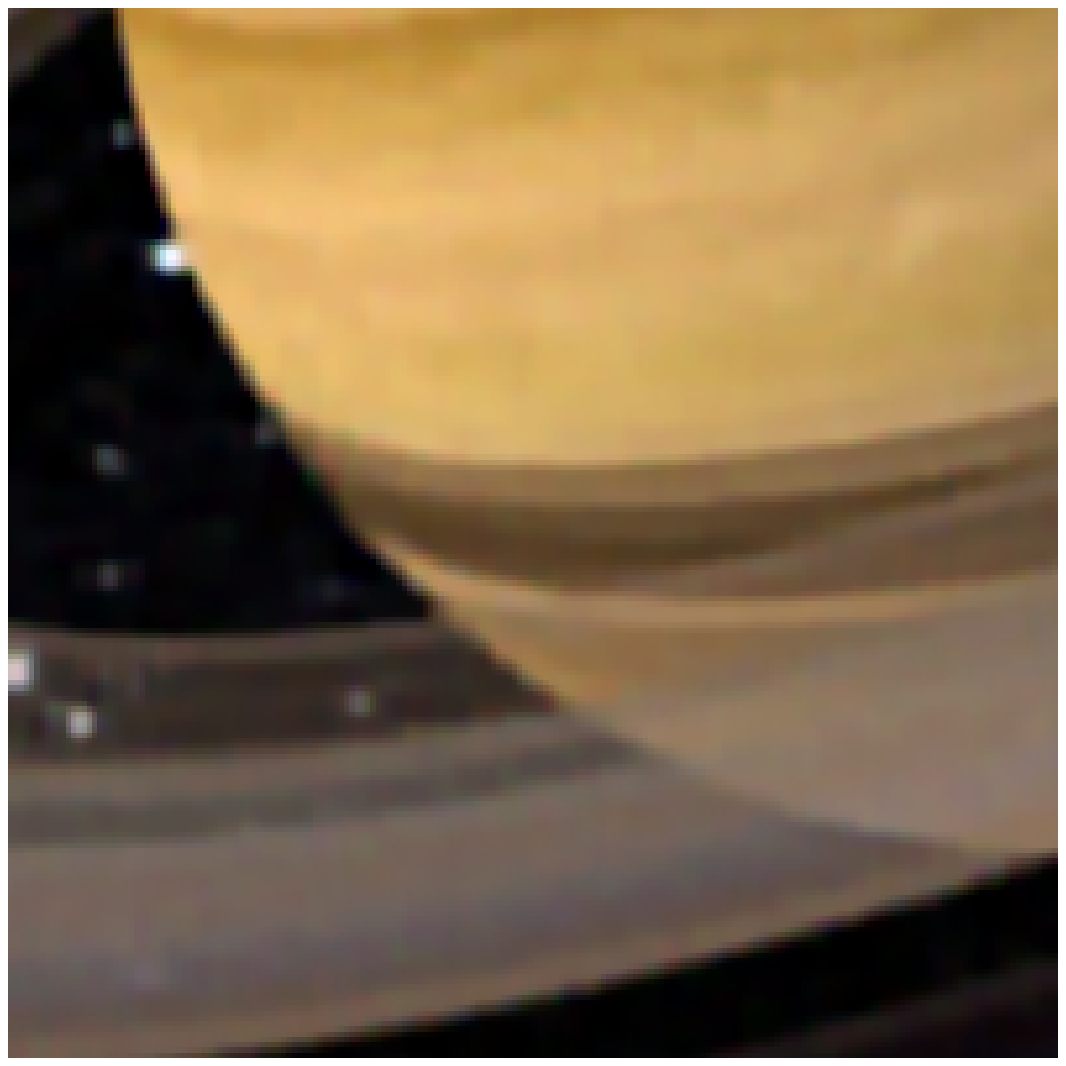}
      \end{minipage}\\
            \begin{minipage}{.185\linewidth}
            \includegraphics[width=\linewidth]{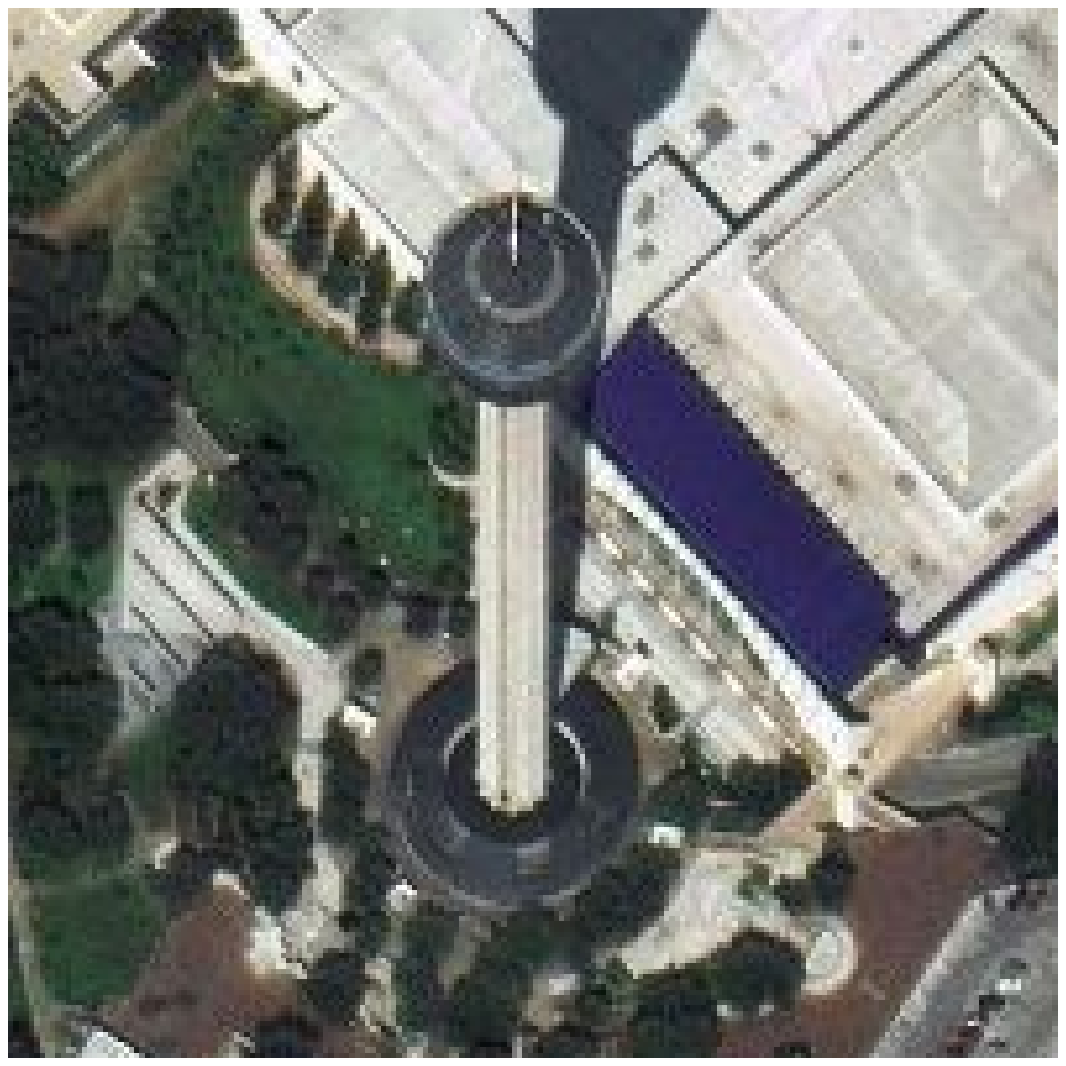}
            \centerline{{(a)}}
      \end{minipage}
      \begin{minipage}{.185\linewidth}
            \includegraphics[width=\linewidth]{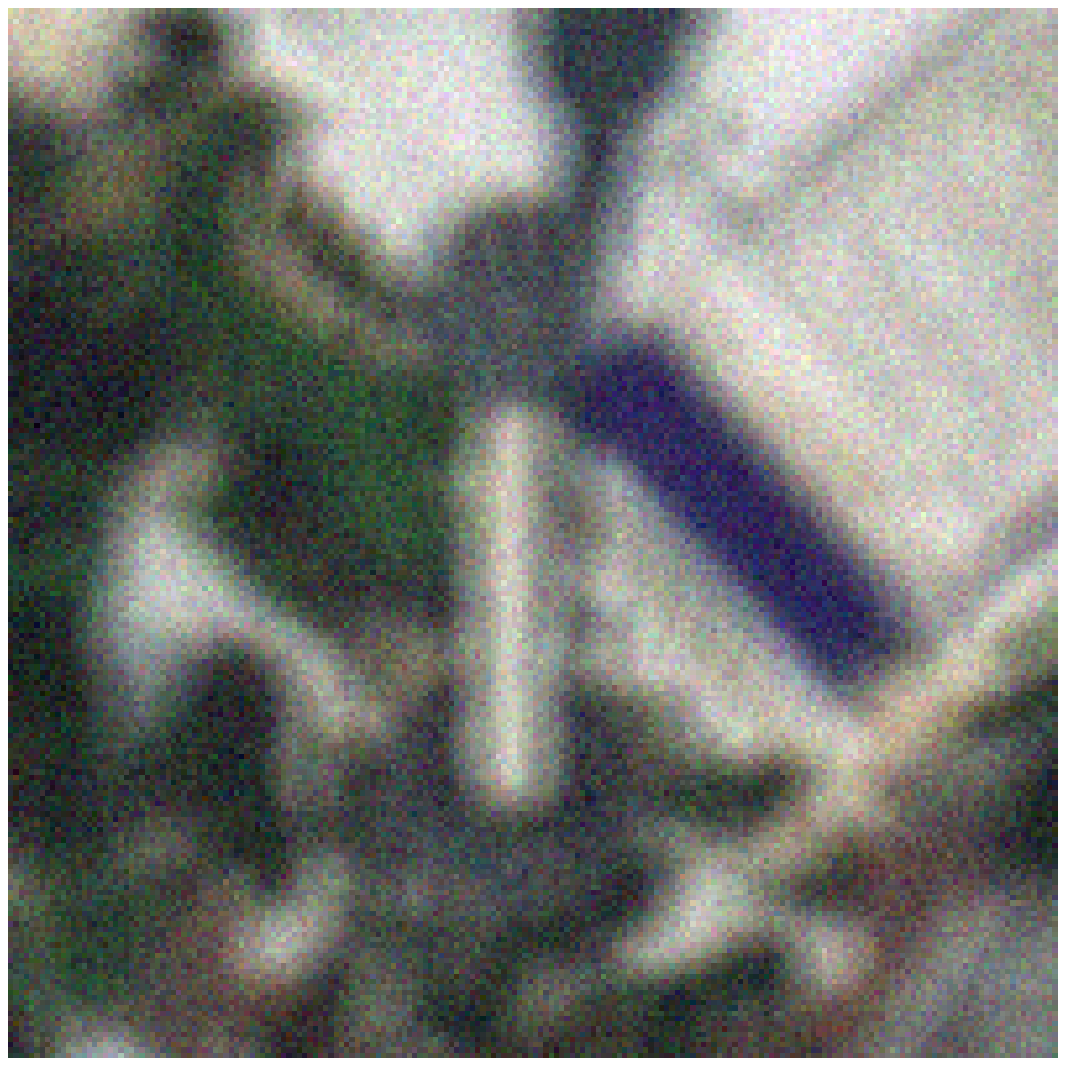}
            \centerline{{(b)}}
      \end{minipage}
       \begin{minipage}{.185\linewidth}
            \includegraphics[width=\linewidth]{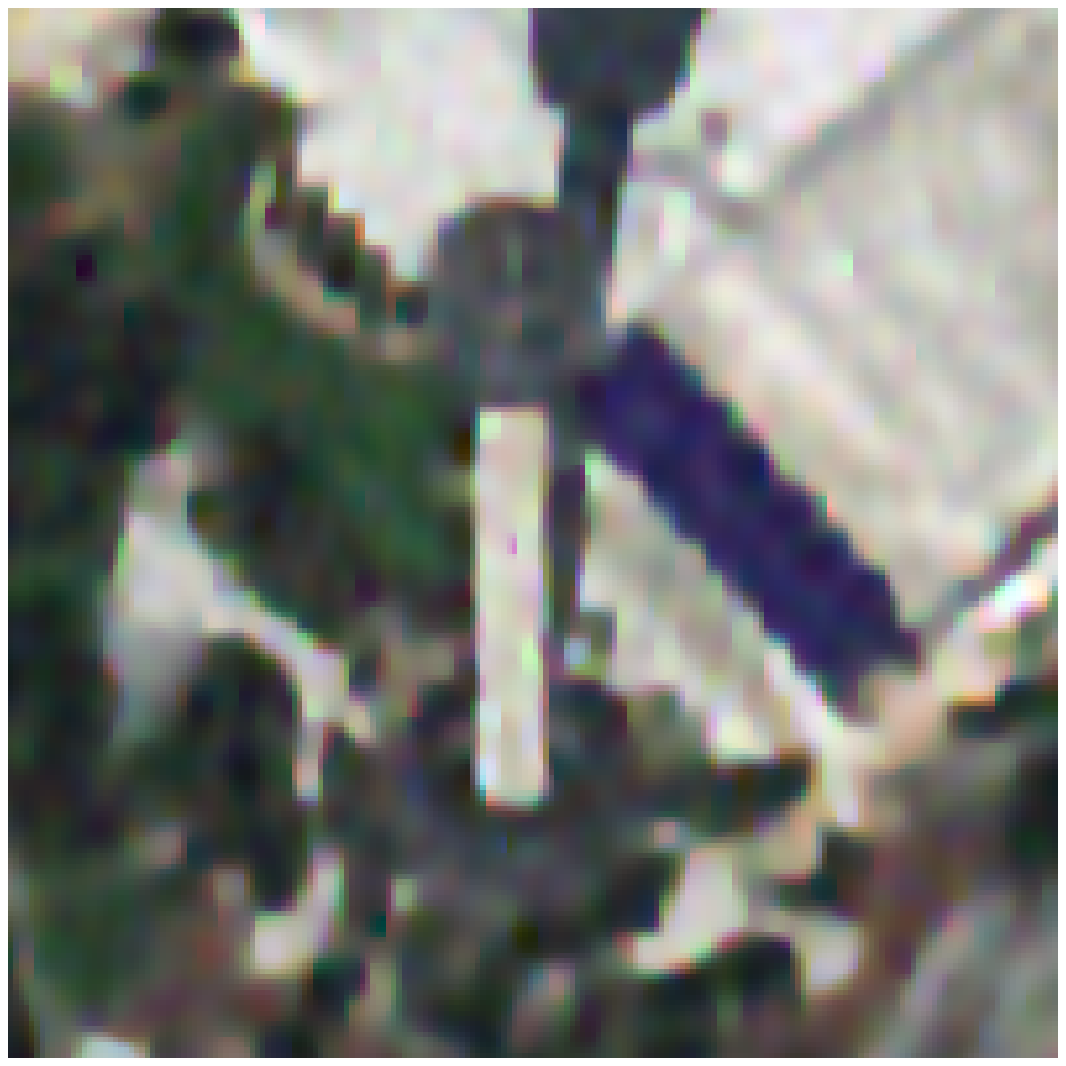}
            \centerline{{(c)}}
      \end{minipage}
      \begin{minipage}{.185\linewidth}
            \includegraphics[width=\linewidth]{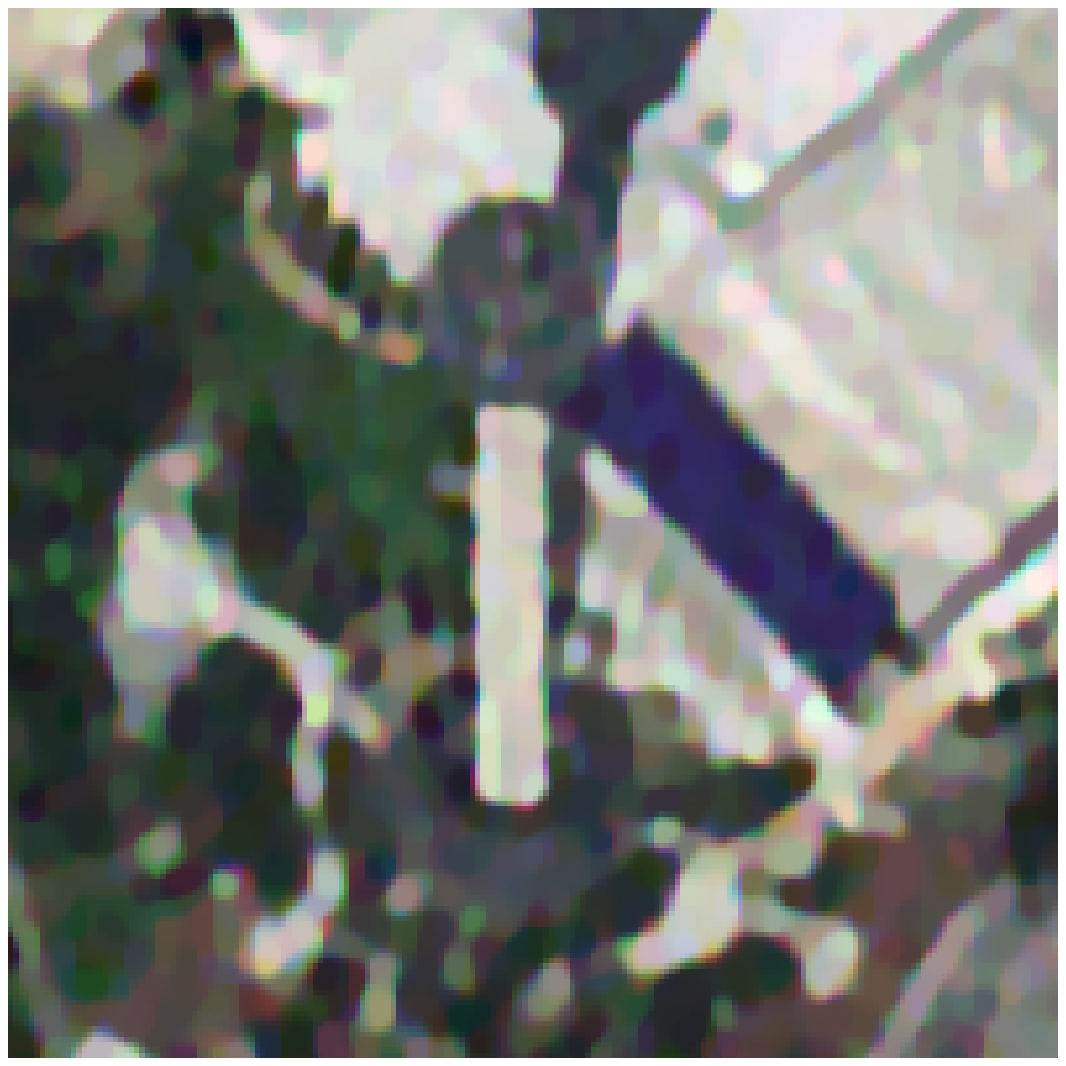}
            \centerline{{(d)}}
      \end{minipage}
      \begin{minipage}{.185\linewidth}
            \includegraphics[width=\linewidth]{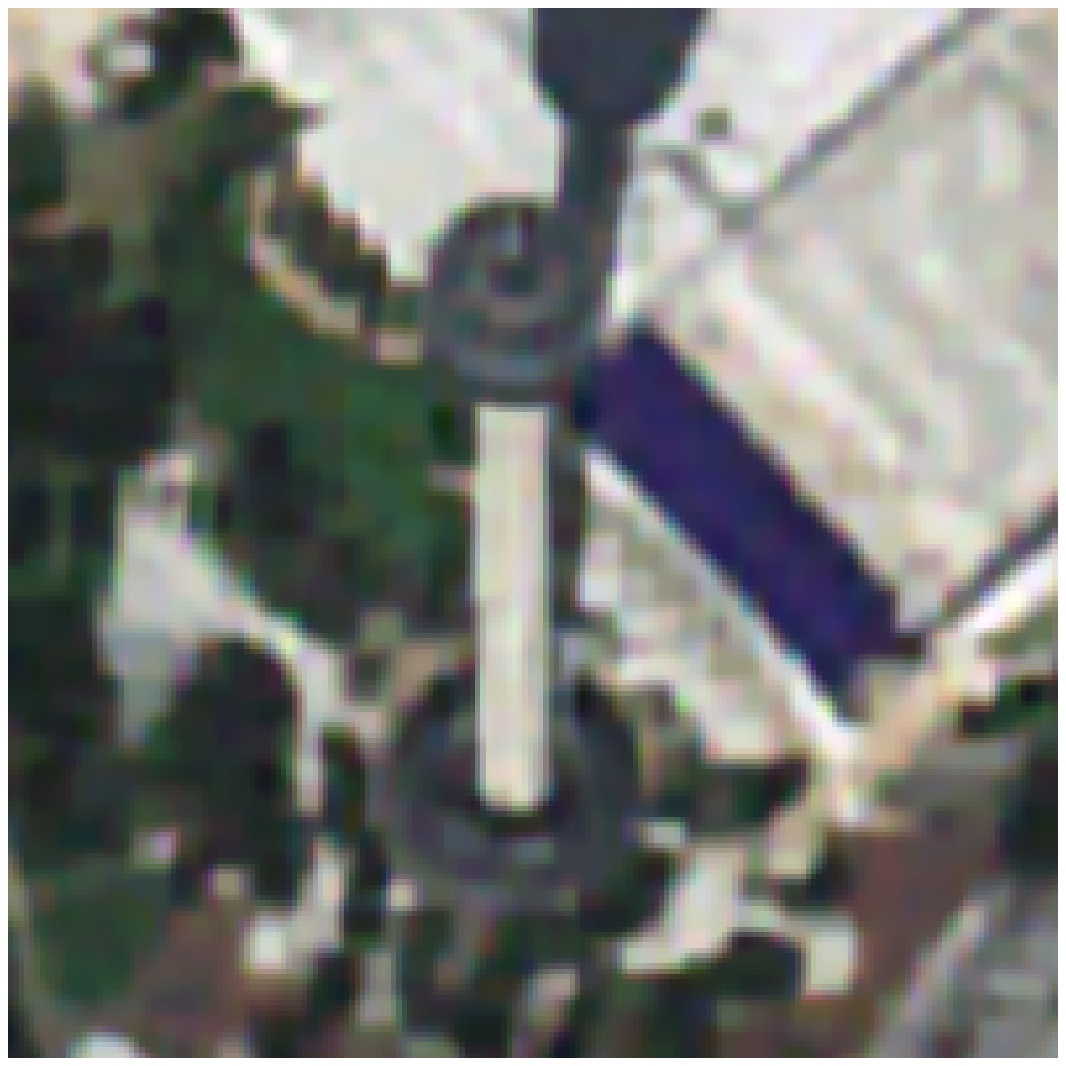}
            \centerline{{(e)}}
      \end{minipage}\\
      \caption{Comparison of magnified local views extracted from deblurring images with spatially-invariant PSF $\sharp 3$. The degraded images are also corrupted by additive Gaussian noise with different levels (i.e., $1\%$, $2\%$ and $5\%$, respectively). From left to right, local views of the (a) original image, (b) blurry + noisy image, deblurred versions of (c) Krishnan's method \cite{Krishnan}, (d) Chan's method \cite{ChanKhoshabeh} and (e) our CNCHTV model are magnified and displayed.}
      \label{noisecase}
\end{figure}

We then focus on the deblurring problem with additive Gaussian noise. All the three images are degraded with the more complex PSF $\sharp3$. For each case, the blurry images are further corrupted by Gaussian noise with different levels (i.e., $1\%$, $2\%$ and $5\%$, respectively). Let $\delta$ denote the noise standard deviation, we tuned and set $\mu = {5}/{\delta}\times 10^{3}$ and $\beta_{i} = 5 \times 10^{2} (i=1,2,3)$ as they provided satisfactory performance. The deblurring results shown in magnified views in Fig.\ref{noisecase} still demonstrate the superior performance of our proposed method. In particular, the other two methods are sensitive to noise and destroy fine image details. In contrast, our proposed scheme can significantly improve the visual deblurring quality. More detailed results in Table 1 show consistently superior performance of our deblurring scheme.
\section{Conclusion}
In this paper, we develop a new constrained hybrid variational deblurring model by combining the non-convex first- and second-order total variation regularizers. To guarantee the high-quality restoration performance, an iteratively reweighted algorithm is proposed based on ADMM. Experimental results show that our proposed method outperforms two existing state-of-the-art deblurring approaches in terms of image details preservation and ringing artifacts suppression.
%

\end{document}